\documentclass[10pt,journal,compsoc]{IEEEtran}
\usepackage{amsmath,amsfonts}
\usepackage{algorithmic}
\usepackage{algorithm}
\usepackage{array}
\usepackage{textcomp}
\usepackage{stfloats}
\usepackage{url}
\usepackage{verbatim}
\usepackage{graphicx}
\usepackage{cite}
\usepackage{color}
\usepackage{multirow}
\usepackage{bbm}
\usepackage{cases}
\usepackage{tabularray}
\usepackage[normalem]{ulem}
\useunder{\uline}{\ul}{}
\usepackage{booktabs}
\usepackage{braket}
\usepackage{amssymb}
\usepackage{float}
\usepackage{threeparttable}
\usepackage{caption}
\usepackage{subcaption}
\usepackage{extarrows}

\providecommand{\zxhreqeq}[1]{Eq.~(\ref{#1})}

\newtheorem{theorem}{Theorem}

\begin{document}

\title{InDeed: Interpretable image deep decomposition with guaranteed generalizability}

\author{Sihan~Wang, Shangqi~Gao, Fuping~Wu, Xiahai~Zhuang$^*$
        % <-this % stops a space
\IEEEcompsocitemizethanks{\IEEEcompsocthanksitem S. Wang and X. Zhuang are with the School of Data Science, Fudan University, Shanghai, China.
\IEEEcompsocthanksitem S. Gao is with the Department of Oncology, University of Cambridge.
\IEEEcompsocthanksitem F. Wu is with the Nuffield Department of Population Health, Oxford University.
\IEEEcompsocthanksitem X. Zhuang$^*$ is the corresponding author. 
E-mail: zxh@fudan.edu.cn}% <-this % stops a space
% \thanks{Manuscript received April 19, 2005; revised August 26, 2015.}
}

\markboth{IEEE TRANSACTIONS ON PATTERN ANALYSIS AND MACHINE INTELLIGENCE,~Vol.~XX, No.~XX, August~2021}%
{Shell \MakeLowercase{\textit{et al.}}: A Sample Article Using IEEEtran.cls for IEEE Journals}

\IEEEpubid{0000--0000/00\$00.00~\copyright~2021 IEEE}

% \maketitle

\IEEEtitleabstractindextext{
\begin{abstract}
%Image decomposition, which involves breaking down an image into elementary components, is essential for numerous downstream tasks. \textcolor{red}{tbc.However, deep learning-based methods have rarely explored explicit decomposition with a focus on interpretability and generalizability.} 
Image decomposition aims to analyze an image into elementary components, which is essential for numerous downstream tasks and also by nature provides certain interpretability to the analysis. 
Deep learning can be powerful for such tasks, but surprisingly their combination with a focus on interpretability and generalizability is rarely explored.
In this work, we introduce a novel framework for interpretable deep image decomposition, combining hierarchical Bayesian modeling and deep learning to create an architecture- modularized and model-generalizable deep neural network (DNN). 
The proposed framework includes three steps: (1) hierarchical Bayesian modeling of image decomposition, (2) transforming the inference problem into optimization tasks, and (3) deep inference via a modularized Bayesian DNN. 
We further establish a theoretical connection between the loss function and the generalization error bound, which inspires a new test-time adaptation approach for out-of-distribution scenarios. 
We instantiated the application using two downstream tasks, \textit{i.e.}, image denoising and unsupervised anomaly detection, and the results demonstrated improved generalizability as well as interpretability of our methods. 
The source code will be released upon the acceptance of this paper.
\end{abstract}

}

\maketitle

\section{Introduction}
\IEEEPARstart{I}{mage} decomposition has shown great potential in subsequent tasks, such as image denoising and anomaly detection\cite{ulyanov2018DIP, kim2018structuredecomposition, gandelsman2019doubleDIP, buades2005denoisingreview}. 
There are two commonly seen scenarios of applications. 
One is to extract representative contents, such as the smoothness components, for which the total variation (TV) technique has been widely used \cite{rudin1992TV}.
The other is to decompose images into compositional parts. Typical methods include the robust principal component analysis \cite{candes_robust_2011} and multi-resolution analysis \cite{ouahabi2012MRA}. 
The former models data matrices as the superposition of low-rank and sparse components, which can be solved via convex optimization; the latter typically transforms an image into components at different scales or resolutions.
These methods could be time-consuming when applied to downstream tasks \cite{kim2018structuredecomposition, babacan_sparse_2012}.

Deep learning-based approaches have shown substantial promise, but recent efforts on image decomposition mainly focus on directly simulating traditional algorithms or heuristic decomposition approaches. 
The simulation methods approximate the mapping of a specific method, such as the neighborhood filter \cite{tomasi1998deepfilterbilateral} and TV-related frameworks \cite{xu2012deepstructureTV, xu2024deepTVlow}; 
the typical heuristic approaches include the idea of "internal patch recurrence" \cite{zontak2011internal} and 
the extraction of homogeneous components \cite{gandelsman2019doubleDIP,zhao2020didfuseDIP,lettry2018darnDIP}.
%For example, Gandelsman \textit{et al.} \cite{gandelsman2019doubleDIP} captured components by minimizing the inherent entropy for image dehazing and foreground segmentation.
To date, most deep learning-based algorithms adopt black-box architectures, which present two significant challenges: (i) limited interpretability due to lack of transparency in the image decomposition process; and (ii) poor generalizability in out-of-distribution (OOD) scenarios. Recently, active interpretability, also known as \textit{ad-hoc} interpretability, has gained ample attention, although it remains underrepresented among the literature \cite{interpretable2021survey}.
Unlike \textit{post-hoc} methods, which try to explain the black box, the \textit{ad-hoc} ones actively design the architecture and/or training process to achieve \emph{model-self} interpretability \cite{interpretable2021survey, rudin2019stop}. 
A notable work is the deep unrolling, which creates links between specific iterative algorithms and deep learning \cite{monga2021unrolling}, and is applicable to various tasks, including image super-resolution, blind image deblurring, and image denoising \cite{monga2021unrolling, liu2019deepunrollingTIP, he2022unrolling+generalization}.
For image decomposition, several methods have been developed with architectures guided by filters \cite{zhang2018interpretablefilter} or variational priors \cite{kim2018structuredecomposition, gao2023bayeseg}. 
% For instance, BayeSeg extracted semantic components, namely appearance, and structure, for medical image segmentation \cite{gao2023bayeseg}. 
Although these methods have demonstrated generalizability in downstream tasks \cite{gao2023bayeseg, he2022unrolling+generalization, monga2021unrolling}, the exploration in image decomposition remains limited.

Herein, we propose a new framework for interpretable deep image decomposition and instantiate its application with two tasks, \textit{i.e.}, image denoising, and unsupervised anomaly detection.
The proposed framework combines Bayesian inference and deep learning, and it consists of three steps, \textit{i.e.},
(1) hierarchical Bayesian modeling, (2) transforming the inference into optimizations, and (3) deep inference. 
In the first step, we decompose an image into statistically or semantically meaningful components, such as low-rank, sparsity, and noise, and then further decompose these components into sub-components to accommodate more priors. Such decomposition is applied recursively via a probabilistic graphic model (PGM) with a hierarchical structure, also known as Hierarchical Bayesian Modeling (HBM).
In the second step, we adopt the variational inference for the HBM, namely approximating the posteriors. This step results in two optimization sub-problems, of which one has a closed-form solution.
In the third step, We design an architecture-modularized deep neural network (DNN) to infer the posteriors, with its architecture based on the HBM in the first step and its training strategy guided by the variational inference in the second step.
\emph{This framework ensures improved interpretability and generalizability for the deep image decomposition models.}

The \textbf{interpretability} originates from the integration of HBM and deep learning, resulting in \emph{ad-hoc} design for both the architecture and training strategy. 
The architecture is modularized according to the HBM, incorporating explicit computations and non-linear mappings to infer the posteriors of corresponding variables, providing interpretable intermediate outputs and a self-explanatory architecture. 
Furthermore, the deep framework mimics the optimization process, and the DNN training strategy is guided by two sub-problems derived from variational inference, resulting in a meaningful loss function.

The \textbf{generalizability} benefits from Bayesian learning and hierarchical modeling. 
Firstly, we utilize  PAC-Bayesian Theory to provide a generalization error bound~\cite{mcallester2003pac_first,mbacke2024pac-vae}, demonstrating that minimization of the loss function aligns with minimization of this error bound. 
Secondly, the hierarchical structure enhances generalizability by fostering the interdependence between variables~\cite{gao2023bayeseg}, which facilitates a sample-specific prior for meaningful components. 
Furthermore, inspired by the error bound and the modularized architecture, we then propose a test-time adaptation algorithm for OOD scenarios.

 Our contributions are summarized as follows:
\begin{itemize}
    \item We propose a new framework to establish an interpretable deep image decomposition. This framework consists of three steps to integrate hierarchical Bayesian modeling and deep learning.
    \item  We establish the theoretical connection between the objective for model training and the generalization error upper bound.  Inspired by this connection, we further propose a test-time adaptation algorithm, which allows for rapid adjustment of the model for OOD scenarios. 
    \item We develop a modularized architecture driven by decomposition modeling to capture targeted components. For the low-rank component in image decomposition, we propose a tailored network that enables learning-based low-rank estimation while allowing for flexible rank adaptation.
    \item We validate the deep image decomposition framework on two downstream tasks,  \textit{i.e.}, image denoising and unsupervised anomaly detection, and the proposed methods demonstrate superior performance.
\end{itemize}

\section{Related works}
\subsection{Image decomposition}
Image decomposition has gained significant attention, driven by the assumption that underlying data often lies in a low-dimensional subspace \cite{babacan_sparse_2012, zhou2010stable2.1, candes2012exact2.1, candes_robust_2011}, offering strong potential in image analysis. Among these techniques, low-rank estimation is attractive with theoretical advances \cite{peng2020robust2.1, candes_robust_2011}, leading to potential modeling options for numerous applications, such as image denoising, anomaly detection, and face recognition.

PCA demonstrates that the low-rank components of matrices are its principal components \cite{candes2012exact2.1, babacan_sparse_2012}, but it often fails with corrupted images. Robust Principal Component Analysis (Robust PCA) \cite{candes_robust_2011} shows that a corrupted data matrix can be decomposed into a low-rank and a sparse component, with theoretical guarantees. Although it can recover the low-rank component exactly, solving RPCA is NP-hard.
Considerable efforts have been made to address this, including the accelerated proximal gradient method by Lin \textit{et al.} \cite{lin2009APG} and the augmented Lagrange multiplier method \cite{lin2010ALM}, which is considered the state-of-the-art for RPCA. However, tuning parameters for optimal performance remains a challenge. Bayesian methods have been introduced to address this, using techniques like Gibbs sampling \cite{zhou2010nonparametric2.1} and variational inference \cite{babacan_sparse_2012} for posterior inference. Despite these advances, the high computational complexity of these methods limits their real-time application.

Recently, deep learning-based image decomposition methods have emerged. Due to the black-box nature of deep neural networks, developing interpretable deep decomposition models remains challenging. Ulyanov \textit{et al.} \cite{ulyanov2018DIP} introduced the "Deep Image Prior"  to capture low-level statistics from a single image, and Gandelsman \textit{et al.} \cite{gandelsman2019doubleDIP} used multiple generator models to decompose images into basic components. More interpretable methods have since been explored, such as RONet for rank-one decomposition \cite{gao_rank-one_2022}, and BayeSeg proposed the appearance-structure decomposition for medical image segmentation \cite{gao2023bayeseg}.
However, research on interpretability and generalizability in terms of deep image decomposition remains limited,
% of these deep methods remain limited, 
warranting further exploration.

\subsection{Image denoising}
Image-denoising approaches can be broadly classified into two categories: traditional model-based methods \cite{buades2005non-denosing-traditional, elad2006image-denosing-traditional, dabov2007image-denosing-traditional} and deep learning-based methods \cite{hu2021pseudo-denoising-deep-self, zhang2017beyond-denoising-deep-resnet, cheng2021nbnet-denoising-deep-encoder, kokkinos2018deep-denoising-deep-resnet}. 
Traditional methods explicitly model the image prior, such as total variation \cite{kim2016TV} and low-rank estimation \cite{el2020LR-DEN}, and are generally agnostic to the type of noise \cite{SIDD_2018_CVPR}. While these methods demonstrate good generalizability, they often fall short in accurately reconstructing image content. 
Deep learning-based algorithms have shown remarkable promise. Various techniques have been introduced to enhance the capabilities, including residual networks \cite{zhang2017beyond-denoising-deep-resnet, kokkinos2018deep-denoising-deep-resnet}, encoder-decoder structures \cite{cheng2021nbnet-denoising-deep-encoder}, and self-similarity approaches \cite{hu2021pseudo-denoising-deep-self}. 

Recently, methods with active interpretability have been proposed for this task. 
For example, Huy Vu \textit{et al.} proposed a deep unrolling network, which integrates the classical graph total variation \cite{denoising-unrolling}. 
Liu \textit{et al.}  proposed an interpretable model for both image generation and restoration (\textit{e.g.} denoising) by decoupling the conventional single denoising diffusion process into residual diffusion and noise diffusion \cite{Liu_2024_diffusion_CVPR}.
These models are typically trained on Gaussian noise in controlled laboratory settings, which can limit their generalizability.
To address this, previous research has focused on generating more realistic noise for training or exposing the networks to a broader spectrum of noise types. 
Chen \textit{et al.} proposed a self-learning strategy that enhances the generalizability by reconstructing masked random pixels of the input image  \cite{chen2023masked}. 
Despite these advancements, the combination of interpretability and generalizability remains under-explored.

\subsection{Unsupervised anomaly detection}
Recent research on unsupervised anomaly detection (UAD) can be classified into two settings, namely, class-specific UAD \cite{zavrtanik_draem_2021, georgescu2021anomaly, domain_yang2020anomaly,zavrtanik_draem_2021} and unified UAD (UUAD) \cite{uniAD}. The former targets the detection of a specific class of objects with a single model \cite{zavrtanik_draem_2021, georgescu2021anomaly, domain_yang2020anomaly}, and the latter tackles a more challenging but vital issue, that is to develop a unified model for the detection of various classes \cite{uniAD}. 
Class-specific UAD methods mainly focus on the modeling of normality and identifying the outliers as anomalies. Mainstream approaches could be classified into three types, \textit{i.e.}, reconstruction-based \cite{ristea_self-supervised_2022, AE_akccay2019skip, AE_hasan2016learning, zavrtanik_draem_2021, dehaene2020FAVAE,vae_liu2021hybrid, yu2021fastflow}, self-learning-based \cite{cutoff_devries2017improved, li_cutpaste_nodate, li2020superpixel, yan2021learning}, and feature extraction-based methods \cite{feature_bergmann2020uninformed, pang_deep_2022,keller2012hics}. Despite different techniques, these methods attempt to learn a better representation of the in-distribution samples. 
Recently, UUAD has attracted increasing attention, albeit to a limited number of publications \cite{uniAD}. UniAD \cite{uniAD} follows the idea of class-specific UAD and aims to design a powerful architecture to capture various in-distribution representations for outliers detection. However, despite the recent promising performances, further study in terms of interpretable and generalizable class-free anomaly detection is required.

\begin{figure*}
\includegraphics[width=1\linewidth]{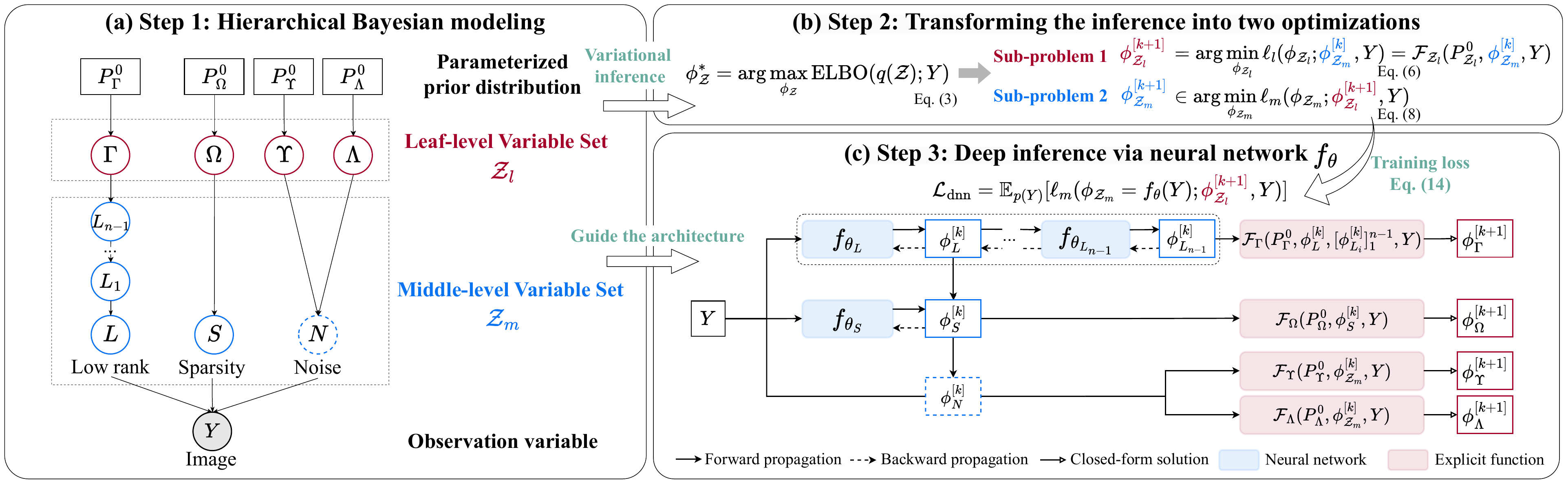}
% Figure/Method/PGM-Structure-v2.pdf
% {Figure/Method/PGM-Structure-v1.pdf}
    \caption{The proposed three-step framework for establishing architecture-modularized and interpretable DNN. Each subfigure illustrates the corresponding step, with Step 3 being developed under the guidance of Steps 1 and 2. Note that the variable $N$ in (a) is marked with a dotted circle, signifying that no inference is required. Function $\mathcal{F}_{X}(\cdot)$ in (b) and (c) denotes closed-form solution(s) w.r.t. variable(s) $X$.
    % The proposed three-step framework for establishing architecture-modularized and interpretable DNN. 
    % % Sub-figure (a) is the PGM of image decomposition and variable $N$ is denoted using a dotted circle because it can be completely determined by $\Upsilon$ and $\Lambda$. 
    % (a) shows the PGM for image decomposition, where variable $N$ is represented by a dotted circle indicating that no inference is required for  $N$. 
    % %\textcolor{blue}{Since $\Upsilon$ and $\Lambda$ can completely determine $N$, it is denoted as a dotted circle. 
    % In (b) and (c), function $\mathcal{F}_{X}(\cdot)$ denotes closed-form solution(s) w.r.t. variable(s) $X$. (c) illustrate $f_{\theta}$, whose structure is guided by (a) and the training loss is conducted by (b).
    }
    \label{fig:summary}
\end{figure*}

\section{Methodology}

In this work, we propose a new framework to develop an architecture-modularized and interpretable deep neural network for image decomposition.  
This framework consists of three steps, and in the following, we first summarize the three steps in Section~\ref{Section3:summary}. Then, we elaborate on each step in sections of~\ref{Section3:modeling},~\ref{Section3: VI} and~\ref{section: deep inference}, respectively. 
In Section~\ref{Method:generalization}, we analyze the generalization error bound and propose an algorithm for test-time adaptation in out-of-distribution (OOD) scenarios. 
Finally, we instantiate two downstream applications,  \textit{i.e.}, image denoising and unsupervised anomaly detection in Section~\ref{Section3: application}.

\begin{table}
\centering
\caption{Summary of mathematical notions and corresponding notations. 
% Herein, a capital letter refers to a matrix, e.g., $A$, a boldface lowercase letter denotes a vector, e.g., $\boldsymbol{a}$, and a lowercase letter refers to a scalar, e.g., \emph{a}. Moreover, a notation superscribed with $0$ denotes a hyperparameter, e.g., $a^0$. 
}
\label{Notation table}
\scalebox{0.78}{
\begin{tabular}{l|l}
\hline
 Notation & Notion \\ \hline
 Lowercase letter &    Scalar (\textit{e.g.}, $a$)  \\
 Bold lowercase letter      &  Vector (\textit{e.g.}, $\boldsymbol{a}$)  \\
 Capital letter  &   Matrix (\textit{e.g.}, $A$)\\
 $\|\cdot\|_2$; $\|\cdot\|_F$; $\Braket{\cdot,\cdot}_F$ & $\ell_2$ norm; Frobenius norm; Frobenius inner product\\ \hline
    $Y \in \mathbb{R}^{h \times w}$      &   Image     \\
  $U \in \mathbb{R}^{h \times w}$&Target (Ground-Truth)\\ \hline
    $L \in \mathbb{R}^{h \times w}$      &   Low rank variable ( $L=AB^T$)\\ 
    $A \in \mathbb{R}^{h \times r^0}$      &   Left factor of low rank component\\ 
    $B^T \in \mathbb{R}^{r^0 \times w}$      &  Right factor of low rank component\\
    $S \in \mathbb{R}^{h \times w}$      &   Sparsity variable\\
    $N \in \mathbb{R}^{h \times w}$      &   Noise variable \\
   \hline
   %  $\boldsymbol{\gamma} \in \mathcal{R}^{r^0},\boldsymbol{\tau} \in \mathcal{R}^{r^0}$&    Variables correlated to $A$ and $B$\\
   %  $\Omega \in \mathcal{R}^{h \times w}$      &    Variable correlated to $S$\\
   %  $ \Lambda \in \mathcal{R}^{h \times w}$    &   Variable correlated to $N$\\ 
   % \hline
$\ell^{(i)}_{\mathrm{X}}; \mathcal{L}(\cdot)/ \hat{\mathcal{L}}(\cdot)$& Loss term of X for sample $i$; expected/ empirical loss\\
$q(\cdot)$, $q_{\phi}(\cdot)$&Variational distribution, parameterized by $\phi$\\
 $\phi_{\mathcal{Z}}$&Variational parameters for variable set $\mathcal{Z}$\\
 $\phi_{\mathcal{Z}}^{[k]}$&Variational parameters for $\mathcal{Z}$ at the $k$-th iteration\\
 $\mu_{X},\sigma_{X}/\ \alpha_Y,\beta_Y$ & Gaussian/ Gamma distribution parameters of variable $X$/ $Y$\\
    % $\psi=\zeta \cup \phi$& Parameter set of prior and variational distributions\\
 $f_\theta(\cdot)$; $f_{\theta_X}(\cdot)$& DNN parameterized by $\theta$; DNN w.r.t. $X$ with $\theta_X$\\
  $\mathcal{F}_{\mathcal{Z}}(\cdot)$& Explicit functions w.r.t. $\mathcal{Z}$\\
   \hline
   $\mathcal{D}$; $|\mathcal{D}|$ & Dataset, the size of a dataset\\
   $h\times w$; $k$ & Image size; iteration index\\
   $x^0$ (\textit{e.g.}, $r^{0})$ & Hyper-parameters (\textit{e.g.}, the maximal rank) \\
   \hline
\end{tabular}}
\end{table}

\subsection{The three-step framework: A summary} \label{Section3:summary}

Fig.~\ref{fig:summary} illustrates the three-step framework. 
In the first step, we employ hierarchical Bayesian modeling (HBM) to describe the procedure of image decomposition and incorporation of prior knowledge. 
This modeling results in a structured graph, which is also referred to as the probabilistic graphical model (PGM). 
In the second step, we use variational inference to approximate the posteriors from the HBM. This step leads to two optimization problems, of which one has closed-form solutions. 
In the third step, we solve the other optimization problem using a modularized Bayesian deep neural network (denoted as $f_{\theta}$), thus this step is referred to as \textit{deep inference}. 
These three steps are described in the sections of~\ref{3.1.1}, \ref{3.1.2} and \ref{Section3.1.3}, respectively. 

Note that the architecture of this neural network ($f_{\theta}$) follows the structure of the PGM in the first step, and its training is guided by the optimization problems in the second step. 
Hence, we refer to $f_{\theta}$ as being architecture-modularized and interpretable.
In Section~\ref{3.1.4}, we analyze the interpretability, modularity, and generalizability of this neural network.

\subsubsection{Modeling of image decomposition}\label{3.1.1}

In the following, we describe the methodological formulations. 
For convenience, we list in Table~\ref{Notation table} the essential mathematical symbols used in this paper.  

Following the conventional decomposition rule \cite{candes_robust_2011}, an image observation $Y=[y_{ij}]_{h \times w}$ can be modeled as a superposition of three components, \textit{i.e.}, the low rank, the sparsity, and the noise, which are denoted respectively by three variables, $L=[l_{ij}]_{h \times w} $, $S = [s_{ij}]_{h \times w}  $, and $N = [n_{ij}]_{h \times w}$. 
Therefore, we have $Y = L+S+N$, and the likelihood function can be written as follows,  
\begin{equation} \label{eq:likelihood}
    p(Y|L,S) = \mathcal{N}\left(Y-(L+S)\ |\ \Upsilon, \Lambda^{-1} \right),
\end{equation}
assuming the noise follows a normal distribution $\mathcal{N}$ with element-wise mean $\Upsilon$ and inverse variance (precision) $\Lambda$.

As the PGM in Fig.~\ref{fig:summary} (a) shows, the observed variable $Y$ is decomposed into three non-observation components,  \textit{i.e.}, $N$, $S$ and $L$, each represented with varying levels of hierarchical structure.
\textbf{For $N$}, no middle-level variable is involved; we directly model the leaf-level variables, \textit{i.e.}, $\Upsilon$ and $\Lambda$, following the likelihood in Eq.~\eqref{eq:likelihood}, with their prior distributions $P_{\Upsilon}^0$ and $P_{\Lambda}^0$. Consequently, $N$ is represented with a dotted circle to indicate that no inference is required.
% there is no middle-level variable, we directly model the leaf-level variables with prior distributions parameterized by $P_{\Upsilon}^0$ and $P_{\Lambda}^0$, based on the likelihood in Eq.~\eqref{eq:likelihood}. Hence, $N$ is represented with a dotted circle, without the need for inference.
% it is represented with a dotted circle
% it is represented with a dotted circle, as it is fully determined by two leaf-level variables, $\Upsilon$ and $\Lambda$. There is no middle-level variable for $N$, and we directly model the leaf-level variables with prior distributions parameterized by $P_{\Upsilon}^0$ and $P_{\Lambda}^0$.}
\textbf{For $S$}, we do not further decompose it, but assign a leaf-level variable ($\Omega$) to model its condition, whose prior distribution is $P_{\Omega}^0$. Consequently, there is only one middle-level variable ($S$ itself) for modeling the sparsity.
\textbf{For $L$}, we first decompose it into a middle-level variable ($L_1$), and then recursively perform such decomposition on the sub-level variables, \textit{i.e.}, from $L_2$ to $L_{n-1}$, until no more sub-component can be applied. This results in a hierarchical structure with $L_{n-1}$ as the final middle-level variable. Similar to $S$, we introduce the condition of $L_{n-1}$ using a leaf-level variable $\Gamma$, whose prior distribution is $P_{\Gamma}^0$. 
As a result, the modeling of $L$ involves $n$ levels of middle-level variables, where $n\ge 2$, along with a leaf-level variable. 
% Hence, one can consider there are $n$, $n\ge 2$, levels of middle-level variables, plus a leaf-level variable, for modeling the low-rank component of $Y$.

% recursively into further middle-level variables (\textit{e.g.} $L_2$) until no additional sub-components are needed, with $L_{n-1}$ as the 
% \textcolor{blue}{\textbf{For $N$}, it is represented with a dotted circle since it is fully determined by two leaf-level variables, $\Upsilon$ and $\Lambda$. Thus, there is no middle-level variable for $N$, and we directly model the leaf-level variables with prior distributions parameterized by $P_{\Upsilon}^0$ and $P_{\Lambda}^0$.}
% \textbf{For $S$}, we do not further decompose it, 
% and thus assign a leaf-level variable ($\Omega$) to model its depending variable, whose prior distribution is parameterized by $P_{\Omega}^0$. 
% Hence, one can consider there is solely one middle-level variable ($S$ itself) for modeling of sparsity.
% \textbf{For $L$}, one could decompose it into new middle-level variable ($L_1$), and further decompose $L_1$ into $L_2$ and likewise recursively until no more sub-component is needed for the final middle-level variable ($L_{n-1}$). 
% Similar to $S$, we then introduce the depending variable of $L_{n-1}$ using a leaf-level variable $\Gamma$, whose prior distribution is parameterized by $P_{\Gamma}^0$. 
% Hence, one can consider there are $n$, $n\ge 2$, levels of middle-level variables, plus a leaf-level variable, for modeling the low-rank component of $Y$.

The above-mentioned three types of levels, \textit{i.e.}, 0, 1 and $n (n\ge 2)$ respectively for $N$, $S$, and $L$, represent the three common scenarios one would encounter for modeling of decomposition.
The details of our image decomposition, modeling, and the assignment of prior distributions for leaf-level variables, will be elaborated on in Section~\ref{Section3:modeling}.

For convenience, we use $\mathcal{Z}$ to denote the set of non-observation variables, representing decomposed components, \textit{i.e.}, $\mathcal{Z}=\{L, L_1, \ldots, L_{n-1}, S, \Gamma, \Omega, \Upsilon, \Lambda\}$.
Furthermore, we divide this variable set into two subsets, \textit{i.e.}, the \textit{\textbf{m}iddle-level} variable set $\mathcal{Z}_m= \{L, L_1, \ldots, L_{n-1}, S\}$ and the \textit{\textbf{l}eaf-level} variable set $\mathcal{Z}_l = \{\Gamma, \Omega, \Upsilon, \Lambda\}$. In the next section, we describe the inference problem of these non-observation variables.

\subsubsection{Formulating inference into two sub-problems}
\label{3.1.2}

The objective of the above hierarchical modeling is to infer the posterior of each variable given an observation, \textit{i.e.}, $p(Z|Y)$ for any $Z\in\mathcal{Z}$. 
Generally, these posteriors are however intractable.
Therefore, we adopt the variational inference approach to approximate them with variational distribution $q_{\phi_{\mathcal Z}}(\mathcal Z|Y)$, parameterized by $\phi_{\mathcal{Z}}$. 
Note that in the following, we abbreviate variational distribution $q_{\phi_{\mathcal{Z}}}(\mathcal{Z}|Y)$ as $q(\mathcal{Z})$ for convenience, unless stated otherwise.

Following the mean-field theory \cite{blei2017VI_survy}, variable groups from different levels can be assumed to be independent, namely, 
\begin{equation}\label{eq:q z} \small
        q(\mathcal Z) = q(\mathcal{Z}_m)q(\mathcal{Z}_l).
\end{equation}

One can obtain variational approximations by minimizing the  Kullback-Leibler (KL) divergence between $q(\mathcal{Z})$ and $p(\mathcal{Z}|Y)$, namely,  $\mathrm{KL}(q(\mathcal{Z} )\|p(\mathcal{Z}|Y))$. 
Since $p(\mathcal{Z}|Y)$ is intractable, we instead equivalently maximize the Evidence Lower BOund (ELBO) of $\log p(Y)$ \cite{zhang2018advancesVI}, namely,
\begin{equation} \label{eq:ELBO} \small
\phi^{*}_{\mathcal{Z}} \in \arg \max_{\phi_{\mathcal{Z}}} \ \mathrm{ELBO}(q(\mathcal{Z}); Y),  
%\log p(Y) - \mathrm{KL}(q(\mathcal{Z}) \| p(\mathcal{Z}|Y)). 感觉这个不重要给个引用，更重要 
\end{equation}
where,
\begin{equation} \label{eq:ELBO function} \small
    \mathrm{ELBO} (q(\mathcal{Z}); Y)= \mathbb{E}_{q(\mathcal{Z})}[\log p(Y|\mathcal{Z})] -\mathrm{KL}(q(\mathcal{Z})\|p(\mathcal{Z})). 
\end{equation}

This ELBO has two parts, \textit{i.e.}, a fidelity term on generating observations, and a KL-divergence term related to priors $p(\mathcal{Z})$. According to the partition of $\mathcal{Z}$, \textit{i.e.}, $\mathcal{Z}=\mathcal{Z}_m \cup \mathcal{Z}_l$, one can further break down the two terms in \zxhreqeq{eq:ELBO function} into more components. For example, the KL term over priors can be rewritten as follows, 
\begin{equation}\label{eq: op objective function} \small
\begin{aligned}
\mathrm{KL}[q(\mathcal{Z})\|p(\mathcal{Z})]=&{\mathbb{E}_{q(\mathcal{Z}_l)}[\mathrm{KL}(q(\mathcal{Z}_m) \|
p(\mathcal{Z}_m | \mathcal{Z}_l))]}
\\
&+ \mathrm{KL}(q(\mathcal{Z}_l)\|p(\mathcal{Z}_l)).
\end{aligned}
\end{equation}
Similarly, we can rewrite the fidelity term for specific tasks, which will be illustrated in Section~\ref{Section3: VI}.

Dividing $\mathcal{Z}$ into two sub-sets leads to an efficient solution for the  optimization problem in Eq.~\eqref{eq:ELBO}:
According to Eq.~\eqref{eq:q z}, one can solve the problem via an alternating algorithm, resulting in two sub-problems, of which one has a closed-form solution, as Fig.~\ref{fig:summary} (b) illustrates.

\setlength{\abovedisplayskip}{20pt} 
\noindent \textbf{Sub-problem 1:} At the $[k+1]^{th}$ step, by fixing $q({\mathcal{Z}_m})$ with $\phi_{\mathcal{Z}_m}^{[k]}$, the ELBO maximization problem of Eq.~\eqref{eq:ELBO} can be converted into a sub-problem with respect to $\phi_{\mathcal{Z}_l}$, namely,
\begin{equation} \label{eq:sub-problem1}
      \phi^{[k+1]}_{\mathcal{Z}_l}=  \arg\min_{\phi_{\mathcal{Z}_l}} \ell_{l}(\phi_{\mathcal{Z}_l}; \phi^{[k]}_{\mathcal{Z}_m},Y) ,
\end{equation}
where,
\begin{equation} \label{eq:Z2} \small
    \begin{aligned}
\ell_l(\phi_{\mathcal{Z}_l}; \phi^{[k]}_{\mathcal{Z}_m},Y) =& -\mathrm{ELBO}(q(\mathcal{Z}|\phi_{\mathcal{Z}_m}=\phi^{[k]}_{\mathcal{Z}_m}); Y) \\ 
=&- \mathbb{E}_{q(\mathcal{Z}_l)}\mathbb{E}_{q^{[k]}(\mathcal{Z}_m)}[\log p(Y|\mathcal{Z})] \\
&+ \mathbb{E}_{q(\mathcal{Z}_l)}[\mathrm{KL}(q^{[k]}(\mathcal{Z}_m) \|
p(\mathcal{Z}_m | \mathcal{Z}_l))] \\
&+ \mathrm{KL}(q(\mathcal{Z}_l)\|p(\mathcal{Z}_l))
+ \mathrm{const}.
    \end{aligned}
\end{equation}
Here, $q^{[k]}(\mathcal{Z}_m) = q_{\phi^{[k]}_{\mathcal{Z}_m}}(\mathcal{Z}_m)$. 
By selecting proper priors for leaf-level variables $\mathcal{Z}_l$, \textit{i.e.}, $P_{ \mathcal{Z}_l}^0$, one can obtain a closed-form solution to this problem, denoted as $\mathcal{F}_{\mathcal{Z}_l}(P^{0}_{\mathcal{Z}_l},\phi^{[k]}_{\mathcal{Z}_m},Y)$. 
This is the reason we use the equality sign in Eq.~\eqref{eq:sub-problem1}.
Section~\ref{Section3: VI} will illustrate the solution in detail. 

\setlength{\abovedisplayskip}{20pt} 
\noindent \textbf{Sub-problem 2:} By fixing $q({\mathcal{Z}_l})$ with $\phi^{[k+1]}_{\mathcal{Z}_l}$, the problem of Eq.~\eqref{eq:ELBO} becomes,
\begin{equation}\label{eq:sub-problem2}
       \phi^{[k+1]}_{\mathcal{Z}_m}\in  \arg\min_{\phi_{\mathcal{Z}_m}} \ell_m(\phi_{\mathcal{Z}_m}; \phi^{[k+1]}_{\mathcal{Z}_l},Y) ,
\end{equation}
where,
\begin{equation} \label{eq:Z1} \small
    \begin{aligned} 
       \ell_m(\phi_{\mathcal{Z}_m}; \phi^{[k+1]}_{\mathcal{Z}_l},Y)=&-\mathrm{ELBO}(q(\mathcal{Z}| \phi_{\mathcal{Z}_l}= \phi_{\mathcal{Z}_l}^{[k+1]}); Y)\\
       =&- \mathbb{E}_{q^{[k+1]}(\mathcal{Z}_l)} \mathbb{E}_{q(\mathcal{Z}_m)}[p(Y|\mathcal{Z})]+ \mathrm{const}\\
       &+\mathbb{E}_{q^{[k+1]}(\mathcal{Z}_l)}[\mathrm{KL}(q(\mathcal{Z}_m) \|
p(\mathcal{Z}_m | \mathcal{Z}_l))] .
    \end{aligned}
\end{equation}
For this problem, conventionally one needs to use an iterative optimization algorithm, such as gradient descent, to minimize the objective for each observation.  
Here, we instead propose to adopt a neural network to learn the mapping from any observation to its corresponding solution. 
\emph{This is the motivation for us to develop the next step, namely, the deep inference framework.}

\subsubsection{Deep inference via neural network 
 $f_\theta$} \label{Section3.1.3}
We introduce a DNN, $f_\theta$, parameterized by $\theta$, to infer $q(\mathcal{Z})$ given any observation $Y$,
\begin{equation} \label{eq:phi and f} 
\phi_{\mathcal{Z} }:=f_\theta(Y) , \forall Y\in \mathcal{Y} .
\end{equation}
% Since the variational posteriors can be achieved via maximization of ELBO in 
The problem in Eq.~\eqref{eq:ELBO} then can be converted to an expected form, as follows,
\begin{subequations} \small
    \label{eq:var14}
    \begin{align}
\phi_{\mathcal{Z}}^* \in & \arg\max_{\phi_{\mathcal{Z}}} \mathrm{ELBO}(q_{\phi_{\mathcal{Z}}}(\mathcal{Z});Y)  \quad \qquad \ \ \label{eq:11a}  \\&\quad \quad \Downarrow \mathrm{(converted\ to)} \notag \\ 
\theta^* \in& \cap_{Y\in \mathcal{Y}} \arg\max_{\theta}\mathrm{ELBO}(q_{f_\theta(Y)}(\mathcal{Z});Y), \label{eq:11b} \\[-1.0ex]
&\ \mathrm{where}\ \phi_{\mathcal{Z}}^* = f_{\theta^*}(Y),   \forall Y\in \mathcal{Y} \ \quad \qquad \notag
\\ &\quad \quad \Downarrow  \mathrm{(relaxed\ to)}  \notag \\
\theta^{*} \in& \arg\min_{\theta} \mathcal{L}_{\mathrm{dnn}} (f_{\theta}; p(Y)), \label{eq:11c} \\[-1.0ex]
&\ \mathrm{where}\ \phi_{\mathcal{Z}}^* \approx f_{\theta^*}(Y),   \forall Y\in \mathcal{Y}.\notag
\end{align}
\end{subequations} 
Here, 
% \textcolor{blue}{we have $\mathcal{L}_{\mathrm{dnn}}(f_{\theta}; p(Y))$ (abbreviated as $\mathcal{L}_{\mathrm{dnn}}$)},
\begin{equation} \small \label{eq: empirical loss of ELBO}
\begin{aligned}
    \mathcal{L}_{\mathrm{dnn}} &= \mathbb{E}_{p(Y)} \left[-\mathrm{ELBO}(q_{f_\theta(Y)}(\mathcal{Z});Y) \right]\\
    &= \mathbb{E}_{p(Y)} \left\{ \mathbb{E}_{q(\mathcal{Z})}[-\log p(Y|\mathcal{Z})] \right\}+\mathbb{E}_{p(Y)}\left[\mathrm{KL}(q(\mathcal{Z})\|p(\mathcal{Z}))\right],
\end{aligned} 
\end{equation} 
which is the loss function for a DNN, \textit{i.e.}, $f_{\theta}$. The rationale of relaxation from Eq.~\eqref{eq:11b} to \eqref{eq:11c} lies in the fact that $\theta^{*}$ is an optimal solution to the optimization problem in Eq.~\eqref{eq:11c}, but not vice versa.
% Here, we introduce a DNN, \textit{i.e.}, $f_{\theta}$, to find the solution for all $Y$ from $p(Y)$. %For convenience, $\mathcal{L}_{\mathrm{dnn}} (f_{\theta}; p(Y))$ is abbreviated as $\mathcal{L}_{\mathrm{dnn}}$.

\textbf{The architecture of $f_\theta$ following the PGM:}
To achieve such a DNN $f_\theta$, which computes variational (marginal) posteriors given any observation, we propose to adopt an architecture-modularized neural network, following the structure of PGM in the first step. 
As illustrated in Fig.~\ref{fig:summary} (c), $f_{\theta}$ first infers $L$ and $S$ by predicting their distribution parameters $\phi_{\mathcal{Z}_m}^{[k]}$ to accomplish decomposition, and then addresses the variables in $\mathcal{Z}_{l}$. As the distribution parameters of $\mathcal{Z}_{l}$ can be directly computed using a closed-form mapping $\mathcal{F}_{\mathcal{Z}_l}$ given $\phi_{\mathcal{Z}_m}^{[k]}$, we simplify $f_{\theta}$ to estimate only $\phi_{\mathcal{Z}_m}$, thus convert Eq.~\eqref{eq:phi and f} to $\phi_{\mathcal{Z}_m} = f_{\theta}(Y)$.

\textbf{Training $f_\theta$ via the inference optimization problem:}
The training loss of DNN $f_\theta$ is the variational objective in Eq.~\eqref{eq: empirical loss of ELBO}, \textit{i.e.}, the expectation of the negative ELBO in Eq.~\eqref{eq:ELBO}. Hence, the two sub-problem formulations are applicable here, leading to a simplified training loss from sub-problem 2, since sub-problem 1 has a closed-form solution,
\begin{equation} 
    \theta^* \in \arg\min_{\theta} \mathcal{L}_{\mathrm{dnn}} ,
\end{equation}
where the loss is as follows at the $[k+1]^{th}$ iteration,
\begin{equation} \label{eq:deep inference} \small
    \mathcal{L}_{\mathrm{dnn}} = \mathbb{E}_{p(Y)} \left[\ell_{m}(\phi_{\mathcal{Z}_m}=f_{\theta}(Y); \phi_{\mathcal{Z}_l}^{[k+1]},Y)\right].  
\end{equation}
Here $\phi^{[k+1]}_{\mathcal{Z}_l} $ upon a sample $Y$ is computed using the closed-form solution to Eq.~\eqref{eq:Z2}, namely, $\phi^{[k+1]}_{\mathcal{Z}_l}=\mathcal{F}_{\mathcal{Z}_l}( P_{\mathcal{Z}_l}^0,\phi^{[k]}_{\mathcal{Z}_m}, Y)$, 
where $\phi_{\mathcal{Z}_m}^{[k]}$ are computed via the forward inference with the current DNN parameter $\theta^{[k]}$, \textit{i.e.}, $\phi_{\mathcal{Z}_m}^{[k]}=f_{\theta^{[k]}}(Y)$ for a sample $Y$.
Once computed, detached $\phi^{[k+1]}_{\mathcal{Z}_l}$ is used for the calculation of the loss function Eq.~\eqref{eq:deep inference} to update $\theta$. 
Hence, for the current batch $\mathcal{D}_{B}=\{Y^{(i)}\}^{|\mathcal{D}_B|}_{i=1}$ at the $[k+1]^{th}$ iteration, the training loss of DNN $f_{\theta}$ becomes, 
\begin{equation} \label{eq:loss dnn 3.1.3}
    \begin{aligned}
        \hat{\mathcal{L}}_{\mathrm{dnn}} = & \textstyle \frac{1}{|\mathcal{D}_B|}\sum_{i=1}^{|\mathcal{D}_B|} [\ell_{m}(f_{\theta}(Y^{(i)});\phi_{\mathcal{Z}_l}^{(i)^{[k+1]}}, Y^{(i)})].
    \end{aligned}
\end{equation}
where $\phi_{\mathcal{Z}_l}^{(i)^{[k+1]}}=\mathcal{F}_{\mathcal{Z}_l}(P^{0}_{\mathcal{Z}_l},f_{\theta^{[k]}}(Y^{(i)}), Y^{(i)})$.
\emph{This loss leads to strong generalizability, which will be elaborated on in the following interpretation.}

\subsubsection{Interpretation} \label{3.1.4}

\textbf{Interpretability and modularity of $f_\theta$:}
In this DNN, each module corresponds to a decomposition step from the hierarchical model, with inputs, outputs, and functions aligned with the HBM. The loss function $\mathcal{L}_{\mathrm{dnn}}$ captures the dependencies between modules through output computation, offering clear interpretability.  Section~\ref{Section3:modeling} to~\ref{section: deep inference} detail this interpretation using an image decomposition task.
% In this DNN, each module corresponds to a decomposition step defined during hierarchical modeling.
% The function of each module (decomposition), as well as the inputs and outputs, are consistent with the PGM.
% Furthermore, the loss function $\mathcal{L}_{\mathrm{dnn}}$ reflects the connections and dependence of each module via the joint computation of their outputs and is easy to interpret. 
% In Section~\ref{Section3:modeling} to~\ref{section: deep inference}, we provide details of such interpretation using the example of the image decomposition task. 

\textbf{Generalizability of $f_\theta$:}
The introduction of hierarchical Bayesian modeling, which incorporates prior knowledge of variables and their interactions, establishes assumptions (preferences) about the solutions for the algorithm. This approach encourages the algorithm to prioritize certain solutions and is a potential method to achieve generalizability with a finite training set, as the no-free-lunch theorem says \cite{wolpert1997no,baxter2000nomodel,goyal2022noinductive}.
Below are two concrete explanations.

Firstly, the framework meets the PAC-Bayesian theory, which naturally guarantees a generalization error bound for $f_{\theta}$ \cite{mcallester2003pac_first, mbacke2024pac-vae}, since the design of $f_{\theta}$ is conducted by HBM. 
We will provide a detailed illustration of this in Section~\ref{Method:generalization}. 

Secondly, one can see that the KL-divergence term in Eq.~\eqref{eq:Z1}, \textit{i.e.},
$\mathbb{E}_{q^{[k+1]}(\mathcal{Z}_l)}[\mathrm{KL}(q(\mathcal{Z}_m) \|
p(\mathcal{Z}_m | \mathcal{Z}_l))]$,
can be treated as a \emph{stochastic regularization term}. Specifically, the KL divergence is stochastically disturbed by $q^{[k+1]}(\mathcal{Z}_l)$ computed in forward-propagation from the current DNN at $[k+1]^{\mathrm{th}}$ iteration. Different from a common prior for $\mathcal{Z}_m$, \textit{e.g.} $p(\mathcal{Z}_m)=\mathcal{N}(0,1)$, the hierarchical modeling, $p(\mathcal{Z}_m|\mathcal{Z}_l)$ introduces the stochastic optimization with randomly turbulent objective functions during training, which helps the training procedure escape from local optima and consequently enhances the generalizability of the resulting DNN.

\subsection{Hierarchical Bayesian modeling of decomposition} \label{Section3:modeling}
In this section, we elaborate on the modeling details. We specify the PGM and model each variable in $\mathcal{Z}$ with dedicated priors. Note that we may need to introduce new variables to the PGM or need to prune it, resulting in a new PGM in Fig.~\ref{fig: network structure} (a).
% decomposition of $L$ formulate a specific PGM.
% and model each variable in $\mathcal{Z}$ with dedicated priors. 
% In this section, we model each variable in $\mathcal{Z}$ with dedicated priors. 
% Note that we may introduce new variables to the PGM or prune it for the specific task.
For convenience, we illustrate the modeling of these variables according to separate branches in Fig.~\ref{fig:summary} (a), namely, (1) variables $\{L,\ldots, \Gamma\}$ related to the low-rank modeling, (2) variables $\{S, \Omega$\} related to the sparsity modeling, and (3) variables $\{N, \Upsilon,\Lambda\}$ related to noise.
Particularly, conjugate priors are adopted to ensure the variational posteriors belong to the same distribution family as the priors.

\textbf{To model the low-rank variable ($L$)},
% ensure the variable $L$ exhibits the low-rank property,
we decompose it into two middle-level variables, \textit{i.e.},
the left factor (denoted as $A$) and the right factor (denoted as $B^T$).
Let $A = [\boldsymbol{a}_1, ..., \boldsymbol{a}_{r^0}]$ and $B^T = [\boldsymbol{b}_1, ..., \boldsymbol{b}_{r^0}]^T$, where vectors $\boldsymbol{a}_i \in \mathbb{R}^{h}$, $\boldsymbol{b}_i \in \mathbb{R}^{w}$ and $r^0$ is a hyper-parameter; we then rewrite $L$, as follows,
\begin{equation} \small
    L = AB^T = \textstyle\sum^{r^0}_{i=1} \boldsymbol{a}_i \times \boldsymbol{b}_i^T = \sum_{i=1}^{r^0}O_i. \label{eq: expression of L}
\end{equation}
Hence, the rank of $A$ or $B$ is no more than $r^0$, namely $\mathrm{rank}(A)\le r^0$, $\mathrm{rank}(B)\le r^0$, and $\mathrm{rank}(L)\le r^0$.
The rank of the matrix $O_i=(\boldsymbol{a}_i \times \boldsymbol{b}_i^T)$ is equal to 1 if and only if both $\boldsymbol{a}_i$ and $\boldsymbol{b}_i$ are non-zero vectors. Hence, the rank of $L$ is bounded by the number of rank-one matrices ($\{O_i\}$). 
Therefore, one can achieve the low-rank constraint by controlling the number of non-zero vectors in $A$ and $B$. Specifically, we introduce a leaf-level variable $\boldsymbol{\gamma} \in \mathbb{R}^{r^0}$ to control $A$ and $B$ simultaneously for compact modeling \cite{babacan_sparse_2012}, as Fig.~\ref{fig: network structure} (a) shows, namely,
\begin{align} 
     p(A|\boldsymbol{\gamma}) = \textstyle\prod^{r^0}_{i=1}\mathcal{N}(\boldsymbol{a}_i|0, \gamma_i^{-1}I_{h}),\label{eq:A modeling} \\
     p(B|\boldsymbol{\gamma}) = \textstyle\prod^{r^0}_{i=1}\mathcal{N}(\boldsymbol{b}_i|0, \gamma_i^{-1}I_{w}),\label{eq:B modeling}
\end{align}
where $I_{h}$ refers to the $h \times h$ identity matrix. One can see that if the $i$-th element of $\boldsymbol{\gamma}$, namely $\gamma_i$, tends to be infinity, the controlled columns $\boldsymbol{a}_i$ and $\boldsymbol{b}_i$ tend to be zero vectors and $O_i$ tends to be a zero matrix, resulting in a lower rank of $L$. Hence, the rank of $L$ can be regularized by $\boldsymbol{\gamma}$. 
We further assign a conjugate prior to $\boldsymbol{\gamma}$, as follows,
\begin{equation} \label{eq:model gamma}
    p(\boldsymbol{\gamma}| \alpha_{\gamma}^0, \beta_{\gamma}^0) = \textstyle \prod_{i=1}^{r^0} \mathcal{G}(\gamma_i | \alpha^0_{\gamma},  \beta^0_{\gamma}),
\end{equation}
where, $\mathcal{G}(\cdot, \cdot)$ denotes Gamma distribution, with hyper-parameters $\alpha_{\gamma}^0$ and $\beta_{\gamma}^0\in \mathbb{R}$. 
\emph{Note that, under the above modeling, the variable $L$ is deterministically dependent on $A$ and $B$, namely, $p(L=A\times B^T|A,B)=1$.} Hence, we denote it with a dotted circle in the PGM of Fig.~\ref{fig: network structure}(a), meaning that it can be pruned.

\textbf{To ensure the sparsity of $S$}, we utilize $\Omega \in \mathbb{R}^{h \times w}$ to independently model the elements of $S$, as follows,
\begin{equation} \label{eq: sparse modeling}
    p(S|\Omega)= \textstyle \prod^{h, w}_{i=1,j=1}\mathcal{N}(s_{ij}|0, \omega_{ij}^{-1}).
\end{equation}
One can see that when $\omega_{ij}$ tends to infinity, the corresponding element $s_{ij}$ statistically approaches zero as well. Moreover, we impose a conjugate prior on $\Omega$,
\begin{equation}  \label{eq:model omega}
    p(\Omega|\alpha^0_\omega, \beta^0_\omega) 
    = \textstyle \prod^{h,w}_{i=1, j=1}\mathcal{G}(\omega_{ij} | \alpha_{\omega}^0, \beta_{\omega}^0),
\end{equation}
where $\alpha_{\omega}^0$ and $\beta_{\omega}^0\in \mathbb{R}$ are hyper-parameters. Note that the marginal distribution $p(S) = \int p(S|\Omega) p(\Omega) d\Omega$ is a Student's \textit{t}-distribution, which has been widely utilized to model sparsity \cite{gao_bayesian_2022}.

\textbf{To model the noise ($N$)}, we assume that the noise follows a pixel-wise normal distribution with mean 0 and the inverse variance $\lambda_{ij}$. Here, we use a matrix $\Lambda=[\lambda_{ij}]_{h \times w}$, \textit{referred to as pixel-wise precision matrix}, to denote all $\lambda_{ij}$, thus $p(N|\Lambda)=\prod_{i=1,j=1}^{h\times w} \mathcal{N}(n_{ij}|0,\lambda_{ij}^{-1})$.
To model $\Lambda$, we impose a conjugate prior,
\begin{equation}  \label{eq:model lambda}
    p(\Lambda|\alpha^0_\lambda, \beta^0_\lambda)
    = \textstyle \prod^{h, w}_{i=1,j=1} \mathcal{G}(\lambda_{ij}|\alpha^0_\lambda, \beta^0_\lambda).
\end{equation}
Here, pixel-wise $\lambda_{ij}$ follows the Gamma distribution with a shape parameter $\alpha_{\lambda}^0\in \mathbb{R}$ and an inverse scale parameter $\beta_{\lambda}^0\in \mathbb{R}$.

Moreover, to facilitate a task-specific decomposition, we introduce another observation, namely, the target variable $U = [u_{ij}]_{h \times w}$. Specifically, $U$ refers to the information dependent on the noise-free components. We will instantiate the modeling of $U$ for specific tasks in Section~\ref{Section3: application}.

Fig.~\ref{fig: network structure}~(a) illustrates the compact PGM w.r.t. variables $\mathcal{Z}= \{A,B,S,\boldsymbol{\gamma}, \Omega, \Lambda\}$ given observations $\{ Y , U\}$. Based on this PGM, we adjust the middle-level variable set to $\mathcal{Z}_{m}=\{A,B,S\}$ and the leaf-level variable set to $\mathcal{Z}_{l}=\{\boldsymbol{\gamma}, \Omega, \Lambda\}$.

\begin{table}
\caption{Priors and variational distributions for variables.}
\label{table: Explict variational posterior}
\scalebox{0.8}{
\begin{tabular}{r@{}l|r@{}l}
\hline
  \multicolumn{2}{c}{$p(\mathcal{Z})$}&\multicolumn{2}{|c}{$q(\mathcal{Z})$}\\ \hline
  $p(A|\boldsymbol{\gamma})$& $:= \prod^{r^0}_{i=1}\mathcal{N}(\boldsymbol{a}_i|0, I_h/\gamma_i)$&$q(A)$&$:=\prod^{r^0}_{i=1}\mathcal{N}(\boldsymbol{a}_i|\mu_{\boldsymbol{a}_i}, diag (\sigma^2_{\boldsymbol{a}_i}))$\\
  $p(B|\boldsymbol{\gamma}) $& $:=\prod^{r^0}_{i=1}\mathcal{N}(\boldsymbol{b}_i|0, I_w / \gamma_i)$&$q(B)$&$:=\prod^{r^0}_{i=1}\mathcal{N}(\mathbf b_i|\mu_{\boldsymbol{b}_i}, diag( \sigma^2_{\boldsymbol{b}_i}))$\\
  $p(S|\Omega)$& $:=\prod^{h, w}_{i,j=1}\mathcal{N}(s_{ij}|0, 1/\omega_{ij})$&$q(S|A,B)$&$:=\prod^{h, w}_{i,j=1}\mathcal{N}(s_{ij}|\mu_{s_{ij}}, \sigma^2_{s_{ij}})$\\ \hline
   $p(\boldsymbol{\gamma}| \alpha_{\boldsymbol{\gamma}}^0, \beta_{\boldsymbol{\gamma}}^0)$& $:=\prod_{i=1}^{r^0} \mathcal{G}(\gamma_i | \alpha^0_{\gamma},  \beta^0_{\gamma})$&$q(\boldsymbol{\gamma})$& $:=\prod^{r^0}_{i=1}\mathcal{G}(\gamma_i | \alpha_{\gamma_i}, \beta_{\gamma_i})$\\
   $p(\Omega|\alpha^0_\omega, \beta^0_\omega) $& $:=\prod^{h,w}_{i, j=1}\mathcal{G}(\omega_{ij} | \alpha_{\omega}^0, \beta_{\omega}^0)$&$q(\Omega)$& $ :=\prod^{h,w}_{i,j=1}\mathcal{G}(\omega_{ij} | \alpha_{\omega_{ij}}, \beta_{\omega_{ij}})$\\
   $p(\Lambda|\alpha^0_\lambda, \beta^0_\lambda)$& $:=\prod^{h,w}_{i,j=1} \mathcal{G}(\lambda_{ij}|\alpha^0_\lambda, \beta^0_\lambda)$&$q(\Lambda)$& $:=\prod^{h,w}_{i,j=1}\mathcal{G}(\lambda_{ij} | \alpha_{\lambda_{ij}}, \beta_{\lambda_{ij}})$\\\hline
\end{tabular}}  
\end{table}

\subsection{Convert inference into two optimizations via VI} \label{Section3: VI}
In this section, we elaborate on the details of the optimization problem conducted via variational inference for $\mathcal{Z}$, as Section~\ref{3.1.2} briefs.

Similar to Eq.~\eqref{eq:q z}, we assume the variational distribution $q(\mathcal{Z})$ to be expressed as,
\begin{equation}\label{eq: variational assumption}
    q(\mathcal{Z})= q(A)q(B)q(S|A,B)q(\boldsymbol{\gamma})q(\Omega)q(\Lambda).
\end{equation}
% Given the selected prior above, 
To enable tractable computation of KL-divergence terms in ELBO, we further assume that $q(\mathcal{Z})$ follows the same form of the conjugate priors assigned to $\mathcal{Z}$,
as summarized in Table~\ref{table: Explict variational posterior}.
% , which further enables tractable computation of KL-divergence terms in \eqref{eq:ELBO}. 
Specifically, for each \textbf{middle-level variable} $Z\in\mathcal{Z}_m=\{A,B,S\}$, the variational distribution $q(Z)$ follows the Gaussian distributions with unknown mean-variance parameters $\phi_{Z} = \{\mu_{Z}, \sigma_{Z}\}$, \textit{e.g.},  
$q(A)=\prod^{r^0}_{i=1}\mathcal{N}(\boldsymbol{a}_i|\mu_{\boldsymbol{a}_i}, diag(\sigma^2_{\boldsymbol{a}_i})) $ with parameters $\phi_{A} = \{\mu_{A}, \sigma_{A}\}$, where $\mu_A = [\mu_{\boldsymbol{a}_1},...,\mu_{\boldsymbol{a}_{r^0}}] \in \mathbb{R}^{h \times r^0}$ and $\sigma_A = [\sigma_{\boldsymbol{a}_1},...,\sigma_{\boldsymbol{a}_{r^0}}] \in \mathbb{R}^{h \times r^0}$. Then, we have the parameter set $\phi_{\mathcal{Z}_m} =\{\mu_{A}, \sigma_{A}, \mu_{B}, \sigma_{B}, \mu_{S}, \sigma_{S}\}$. 
Similarly, for each \textbf{leaf-level variable} $Z\in\mathcal{Z}_l = \{\boldsymbol{\gamma}, \Omega, \Lambda\}$, $q(Z)$ follows the Gamma distribution with parameters $\phi_{Z} = \{\alpha_{Z}, \beta_{Z}\}$, \textit{e.g.},  $q(\boldsymbol{\gamma}) = \prod^{r^0}_{i=1}\mathcal{G}(\gamma_i | \alpha_{\gamma_i}, \beta_{\gamma_i})$ with parameters $\phi_{\boldsymbol{\gamma}} = \{\alpha_{\boldsymbol{\gamma}}, \beta_{\boldsymbol{\gamma}}\}$. 
Hence, the variational parameter set for $\mathcal{Z}_l$ is now $\phi_{\mathcal{Z}_l} =  \{ \alpha_{\boldsymbol{\gamma}}, \beta_{\boldsymbol{\gamma}},\alpha_{\Omega}, \beta_{\Omega}, \alpha_\Lambda, \beta_{\Lambda} \}$.

Based on the new PGM in Fig.~\ref{fig: network structure}~(a), when given observations $\{Y,U\}$, the objective of negative ELBO in Eq.~\eqref{eq:ELBO function} can be reformulated as,
\begin{subequations} \label{eq:whole loss function} \small
\begin{align}
    \ell(Y,U)=&-\mathrm{ELBO}(q(\mathcal{Z});Y,U) \notag\\
    =&-\mathbb{E}_{q(\mathcal{Z})}[\log p(Y|\mathcal{Z})] \qquad \quad \    {\scriptstyle {\mathrm{(fidelity \  term:}}\ \ell_{fid})} \\
&- \mathbb{E}_{q(\mathcal{Z}_m)}[\log p(U|\mathcal{Z}_m)] \quad \  {\scriptstyle {\mathrm{(supervision \  term:}}\ \ell_{sup})}\\
    &+\underbrace{\mathbb{E}_{q(\boldsymbol{\gamma})}[\mathrm{KL}(q(A)q(B) \|
p(A|\boldsymbol{\gamma})p(B|\boldsymbol{\gamma}))]}_{{\mathrm{(low-rank\ term}:}\ \ell_{rank})} \\
&+ \underbrace{\mathbb{E}_{q(\Omega)}\mathbb{E}_{q(A,B)}[\mathrm{KL}(q(S|A,B) \|
p(S|\Omega))]}_{{\mathrm{(sparsity\ term:}}\ \ell_{sparse})}\\
&+ \mathrm{KL}(q(\boldsymbol{\gamma}) \|
p(\boldsymbol{\gamma}))  \qquad \qquad   {\scriptstyle(\boldsymbol{\gamma}\mathrm{-prior \  term}:\ \ell_{\boldsymbol{\gamma}})} \label{eq: gamma prior term}\\
&+\mathrm{KL}(q(\Omega)  \|
p(\Omega)) \qquad \quad  \quad {\scriptstyle(\Omega-\mathrm{prior \  term}:\ \ell_{\Omega})} \label{eq: omega prior term}\\
&+\mathrm{KL}(q(\Lambda) \|
p(\Lambda)) \qquad \quad    \quad {\scriptstyle(\Lambda\mathrm{-prior \  term:}\ \ell_{\Lambda})}. \label{eq: lambda prior term}
\end{align}
\end{subequations}
where the terms on the right side are denoted as $\ell_{fid}$, $\ell_{sup}$, $\ell_{rank}$, $\ell_{sparse}$, $\ell_{\boldsymbol{\gamma}}$, $\ell_{\Omega}$, and $\ell_{\Lambda}$, respectively.
Concretely, $\ell_{fid}$ imposes the consistency between the generated data from the posterior distributions and the observation $Y$, and $\ell_{sup}$ drives the generated data to be task-specific targets supervised by the ground truth $U$. Moreover, $\ell_{rank}$ constrains the low-rank property of variable $L$, and $\ell_{sparse}$ enforces variable $S$ to be sparse. Finally,  $\ell_{\boldsymbol{\gamma}}, \ell_{\Omega}, \ \mathrm{and} \ \ell_{\Lambda}$ penalize the distance between variational approximations and priors of their corresponding variables. 
% Finally, $\ell_{\mathrm{fid}}$ imposes the consistency between the generated data from the posterior distributions and the observation $Y$, and $\ell_{\mathrm{sup}}$ drives the generated data to be task-specific targets supervised by the ground truth $U$.

To minimize Eq.~\eqref{eq:whole loss function} w.r.t. $\phi_{\mathcal{Z}}$, we split the objective function and derive two sub-problems, \textit{i.e.}, \textbf{sub-problem 1} w.r.t. the leaf-level variable set $\mathcal{Z}_{l}=\{\boldsymbol{\gamma}, \Omega, \Lambda \}$ and \textbf{sub-problem 2} w.r.t. the middle-level variable set $\mathcal{Z}_{m}=\{A,B,S\}$, as Section~\ref{3.1.2} briefs. In the following, we first illustrate closed-form solutions to sub-problem 1 in Section~\ref{subsection: closed-form z2} and then elaborate on the objective function for sub-problem 2 in Section~\ref{subsection: objective function z1}.

\subsubsection{Closed-form solutions for sub-problem 1} \label{subsection: closed-form z2}
% The closed-form solutions of parameters $\phi_{\mathcal{Z}_l}$ for variables $\mathcal{Z}_l= \{ \boldsymbol{\gamma}, \Omega, \Lambda\}$ can be explicitly derivated by optimizing \eqref{eq: psi_2 objective}. 
At the $[k+1]^{\mathrm{th}}$ iteration, given $\phi^{[k]}_{\mathcal{Z}_m}$ for $q(\mathcal{Z}_m)$ and the specific assignments in Table \ref{table: Explict variational posterior}, one can obtain a closed-form solution $\phi^{[k+1]}_{\mathcal{Z}_l}$ to the problem defined in Eq.~\eqref{eq:sub-problem1}, \textit{i.e.}, minimizing $\ell_{l}(\phi_{\mathcal{Z}_l};\phi_{\mathcal{Z}_m}^{[k]},Y)$ (abbreviated as $\ell_l$), to update $q(\mathcal{Z}_l)$. 
Specifically, after substituting the ELBO in Eq.~\eqref{eq:whole loss function} into Eq.~\eqref{eq:Z2}, $\ell_l$ consists of six terms, \textit{i.e.},  $\ell_{fid}$, $\ell_{rank}$, $\ell_{sparse}$, $\ell_{\boldsymbol{\gamma}}$, $\ell_{\Omega}$, and $\ell_{\Lambda}$ in Eq.~\eqref{eq:whole loss function}, since $\ell_{sup}$ is constant. 
Moreover, due to the independence between $\boldsymbol{\gamma}, \Omega$, and $\Lambda$, their variational distributions can be optimized separately.
In the following, we provide the expressions of these solutions, while
\emph{details of the deduction for their closed-form solutions can be found in Section 1 of Supplementary Material.}
Note that the iteration index denoted with superscript $[k]$ will be omitted here for brevity.

\textbf{For variable $\boldsymbol{\gamma}$}, minimizing $\ell_{l}$ over $q(\boldsymbol{\gamma})$ is equivalent to minimizing
\begin{equation} \small 
    \ell_l (\phi_{\boldsymbol{\gamma}};\phi_A,\phi_B) = \ell_{rank} + \ell_{\boldsymbol{\gamma}},
\end{equation}
which is only related to $\phi_A = \{\mu_{A}, \sigma_{A}\}$ and $\phi_B = \{\mu_{B}, \sigma_{B}\}$. The closed-form solution for each element in $\phi_{\boldsymbol{\gamma}}$, \textit{i.e.}, $\alpha_{\gamma_i}, \beta_{\gamma_i}$, can be derived,
\begin{equation}\label{eq: gamma formula} \small   
\begin{cases}
\alpha_{\boldsymbol{\gamma}_i} = 2\alpha^0_{\boldsymbol{\gamma}} + h + w  \\
\beta_{\boldsymbol{\gamma}_i} = 2\beta^0_{\boldsymbol{\gamma}} + \|\mu_{\boldsymbol{a}_i}\|_2^2+\|\sigma_{\boldsymbol{a}_i}\|_2^2+ \|\mu_{\boldsymbol{b}_i}\|_2^2+\|\sigma_{\boldsymbol{b}_i}\|_2^2.
\end{cases}
\end{equation}
Since $q(\gamma_i)$ is a Gamma distribution, we have $\mu_{\gamma_i}=\mathbb{E}[\gamma_i] = \alpha_{\gamma_i}/\beta_{\gamma_i}$. 
Note that $\mu_{\gamma_i}$ tends to be a larger value when $\beta_{\gamma_i}$ is smaller, which subsequently encourages $a_i$ and $b_i$ to approach zero, resulting in a lower rank for $L$.

\textbf{For variable $\Omega$}, minimizing $\ell_{l}$ over $q(\Omega)$ is equivalent to minimizing
\begin{equation} \small 
    \ell_l (\phi_{\Omega};\phi_S) = \ell_{sparse} + \ell_{\Omega},
\end{equation}
which is only conditioned on $\phi_S = \{ \mu_{S}, \sigma_{S}\}$. Then, the closed-form solution for each element in $\phi_{\Omega}$, \textit{i.e.}, $\alpha_{\omega_{ij}}, \beta_{\omega_{ij}}$, is given by,
\begin{equation}\label{eq: omega formula} \small    
\begin{cases}
\alpha_{\omega_{ij}} = 2 \alpha^0_{\omega} + 1,  \\
\beta_{\omega_{ij}}={2\beta^0_\omega  + \mu_{s_{ij}}^2+ \sigma_{s_{ij}}^2}.
\end{cases}
\end{equation} 
Similarly, we have $\mu_{\omega_{ij}} =\mathbb{E}[\omega_{ij}]= \alpha_{\omega_{ij}}/\beta_{\omega_{ij}}$. When $\beta_{\omega_{ij}}$ is smaller, $\mu_{\omega_{ij}}^{-1}$ tends to drive $s_{ij}$ zero, leading to a sparser $S$.

\textbf{For variable $\Lambda$}, minimizing $\ell_{l}$ over $q(\Lambda)$ is equivalent to minimizing 
\begin{equation} \small 
    \ell_l (\phi_{\Lambda};\phi_{\mathcal{Z}_m}, Y) = \ell_{fid} + \ell_{\Lambda}. 
\end{equation}
The closed-form solution for each element in $\phi_{\Lambda}$, \textit{i.e.}, $\alpha_{\lambda_{ij}}$ and $\beta_{\lambda_{ij}}$, is as follows, 
\begin{equation}\label{eq: lambda formula} \small 
\begin{cases}
\alpha_{\lambda_{ij}}=2\alpha^0_{\lambda} + 1,  \\
\beta_{\lambda_{ij}} =2\beta^0_\lambda + [y_{ij} - (\sum^{r^0}_k \hat{a}_{ik} \times \hat{b}_{jk}+\hat{s}_{ij})]^2,
\end{cases}
\end{equation} 
where $[\hat{a}_{ik}]_{h\times r^0}=\hat{A}, [\hat{b}_{jk}]_{w\times r^0}=\hat{B}$ and $[\hat{s}_{ij}]_{h \times w} = \hat{S}$ are sampled from $q(A), q(B)$ and $q(S|A,B)$, which will be illustrated in the following section. 
Similarly, we have $\mu_{\lambda_{ij}}=\mathbb{E}[\lambda_{ij}] = \alpha_{\lambda_{ij}}/\beta_{\lambda_{ij}}$. The smaller residual error in the denominator leads to the larger value of $\mu_{\lambda_{ij}}$, resulting in a weaker noise level of $n_{ij}$ in $N$.

\subsubsection{The objective function for sub-problem 2} \label{subsection: objective function z1}
By fixing $q(\mathcal{Z}_l)$ with $\phi_{\mathcal{Z}_l}^{[k+1]}$, we need to update $q(\mathcal{Z}_m)$ by minimizing $\ell_m(\phi_{\mathcal{Z}_m}; \phi_{\mathcal{Z}_l}^{[k+1]},Y)$, as defined in Eq.~\eqref{eq:sub-problem2}. After substituting the ELBO in Eq.~\eqref{eq:whole loss function} into Eq.~\eqref{eq:Z1}, the objective function $\ell_m$ w.r.t. $\phi_{\mathcal{Z}_m}$ can be expressed by,
\begin{equation} \label{eq: ob phi_1} \small 
    \ell_m(\phi_{\mathcal{Z}_m}; \phi_{\mathcal{Z}_l},Y,U)= \ell_{fid} + \ell_{sup} 
 + \ell_{rank} + \ell_{sparse} + const,
\end{equation}
since the prior terms related to $\mathcal{Z}_l$, \textit{i.e.}, $\ell_{\gamma}$, $\ell_{\Omega}$, and $\ell_{\Lambda}$ in Eq.~\eqref{eq:whole loss function}, are constant given $q(\mathcal{Z}_l)$.

For the fidelity term  $\ell_{fid}$, it is intractable to compute the expectation directly. Hence, we apply the following equivalent expression and approximate it with the Monte Carlo sampling strategy \cite{doersch2016vae},
% due to the intractability of explicitly computing the expectation, it equals the following expression and is approximated later with the sampling technique,
\begin{equation} \label{eq:l_fid} \small 
\begin{aligned}
    \ell_{fid} &= - \mathbb{E}_{q(\mathcal{Z})} [\log p(Y|\mathcal{Z})] =- \mathbb{E}_{q(\Lambda)} [\mathbb{E}_{q(\mathcal{Z} \backslash \Lambda)} \log p(Y|\mathcal{Z})]\\
    & \approx \frac{1}{2}\left\|\sqrt{\mu_{\Lambda}}\odot (Y - (\hat{A}\hat{B}^T +\hat{S}))\right\|^2_{F}.
\end{aligned} 
\end{equation}
Here, we adopt the reparameterization technique \cite{doersch2016vae}, and have $\hat{A}= \mu_{A} + \sigma_A \odot \eta, \hat{B} = \mu_{B} + \sigma_B \odot \eta$, and $\hat{S} = \mu_{S} + \sigma_S \odot \eta $, where $\eta \sim \mathcal{N}(\mathbf{0},\mathbf{I})$. Moreover, the supervision term $\ell_{sup}$ in Eq.~\eqref{eq:whole loss function} is task-specific which will be instantiated in Section~\ref{Section3: application}.

The low-rank term $\ell_{rank}$ can be expressed as a function w.r.t. parameters of $q(A)$ and $q(B)$, namely, $\phi_A = \{\mu_A, \sigma_A\}$ and $ \phi_B = \{\mu_B, \sigma_B\}$, given the updated expectation of $\boldsymbol{\gamma}$ ($ \mu_{\boldsymbol{\gamma}}=\mathbb{E}[\boldsymbol{\gamma}]= \alpha_{\boldsymbol{\gamma}}/\beta_{\boldsymbol{\gamma}}$ using Eq.~\eqref{eq: gamma formula}),
% given the formulas of conditional distributions $p(A|\boldsymbol{\gamma}), p(B|\boldsymbol{\gamma})$, variational distribu tions $q(A), q(B)$, and $\phi_{\boldsymbol{\gamma}}$, namely,
\begin{equation}  \label{eq: VI L_rank}  \footnotesize
\begin{aligned} 
     \ell_{rank} =& \frac{1}{2}\left\{ \|\mu_A \odot \mathbf{1}_{h}\sqrt{\mu_{\boldsymbol{\gamma}}}^T\|_F^2 +\Braket{\mathbf{1}_{h}\mu_{\boldsymbol{\gamma}}^T, \sigma_A^2}_F -\Braket{\mathbf{1}_{h\times r^0}, \ln \sigma_A^2}_F \right.  \\
     +& \left.\|\mu_B \odot \mathbf{1}_{w}\sqrt{\mu_{\boldsymbol{\gamma}}}^T\|_F^2+\Braket{\mathbf{1}_{w}\mu_{\boldsymbol{\gamma}}^T, \sigma_B^2}_F -\Braket{\mathbf{1}_{w\times r^0}, \ln \sigma_B^2}_F\right\},  
\end{aligned} 
\end{equation}
where $\odot$ refers to Hadamard product and $\mathbf{1}_{x}$ is a vector or matrix of ones with shape $x$.

Similarly, the sparsity term $\ell_{sparse}$ can be expressed as a function w.r.t. parameters of $q(S)$, namely $\phi_{S} = \{\mu_S,\sigma_S\}$, given the updated expectation of $\Omega$ ($\mu_{\Omega} = \mathbb{E}[\Omega]=\alpha_{\Omega}/\beta_{\Omega}$ using Eq.~\eqref{eq: omega formula}),
\begin{equation}\label{eq:VI L_sparse} \small
     \ell_{sparse}
     =\frac{1}{2}\left\{\|\mu_S \odot \sqrt{\mu_\Omega}\|^2_F+\Braket{\mu_{\Omega}, \sigma_S^2}_F - \Braket{\textbf{1}_{h\times w}, \ln\sigma_S^2}_F\right\}.
\end{equation} 

As Section~\ref{Section3.1.3} briefs, we propose to build a neural network to infer the solutions instead of individually optimizing $\phi_{\mathcal{Z}_m}$ for each observation, which will be illustrated in the next section.

\begin{figure*}%[!b]
    \centering
\includegraphics[width=0.95\linewidth]{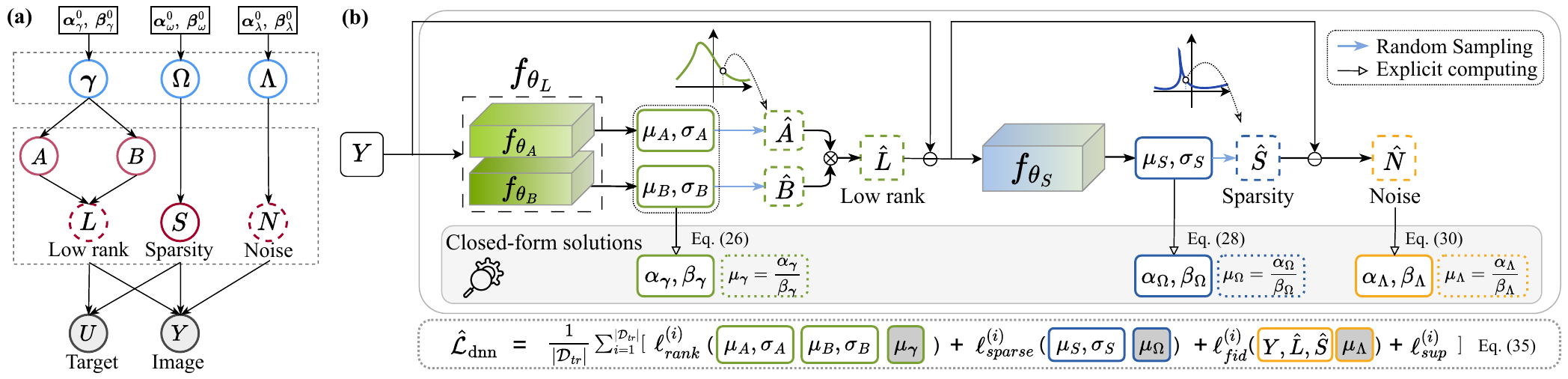}
% {Figure/Method/LRNet-Structure1.pdf}Network-Structure
   \caption{The PGM and architecture of InDeed. (a) illustrates the PGM with observation variables $\{Y,U\}$ and non-observation $\mathcal{Z}= \{A,B,S,\boldsymbol{\gamma}, \Omega, \Lambda \}$. No inference is required for variables $L$ and $N$ in dotted circles.  
    (b) shows the architecture. Given an observation $Y$, the posteriors of middle-level variables, \textit{i.e.}, $A, B$ and $S$, are first inferred via neural networks, and then the expectations of leaf-level variables, \textit{i.e.}, $\mu_{\boldsymbol{\gamma}}, \mu_{\Omega}$ and $\mu_{\Lambda}$, are given by the closed-form solutions in Eq.~\eqref{eq: gamma formula}, Eq.~\eqref{eq: omega formula}, and Eq.~\eqref{eq: lambda formula}, respectively.  The shaded parameters are detached values during training.} 
    \label{fig: network structure}
\end{figure*}

\subsection{InDeed: Deep inference}
\label{section: deep inference}
In this section, we elaborate on details of the proposed \textbf{in}terpretable \textbf{dee}p \textbf{d}ecomposition inference method, \textit{i.e.,} InDeed, including the objective function, the architecture, and the interpretation. %ble insights.

\subsubsection{Objective function}
Given a training data set $\mathcal{D}_{tr} = \{Y^{(i)}, U^{(i)}\}^{|\mathcal{D}_{tr}|}_{i=1} $, the objective function of $f_{\theta}$, namely, the \emph{empirical} version of Eq.~\eqref{eq:deep inference} can be expressed as, combining Eq.~\eqref{eq: ob phi_1},
% formulated in Eq.~\eqref{eq:loss dnn 3.1.3} now can be unfolded after substituting Eq.~\eqref{eq: empirical loss of ELBO}\textcolor{red}{(wfp: no substituting of (12)?)} and expectation of Eq.~\eqref{eq: ob phi_1} using \emph{empirical} training loss,  as follows,
\begin{equation} \label{eq: deep loss function} \small
    \begin{aligned}
        \hat{\mathcal{L}}_{\mathrm{dnn}} &= \frac{1}{ |\mathcal{D}_{tr}|}\textstyle \sum_{i=1}^{|\mathcal{D}_{tr}|} \left[\ell_{m}\left (\phi_{\mathcal{Z}_m}=f_{\theta}(Y^{(i)}); \phi_{\mathcal{Z}_l}^{[k+1]},Y^{(i)},U^{(i)}\right)\right]\\
        &= \frac{1}{|\mathcal{D}_{tr}|}\textstyle \sum_{i=1}^{|\mathcal{D}_{tr}|}[ \ell^{(i)}_{fid} + \ell^{(i)}_{sup} + \ell^{(i)}_{rank} + \ell^{(i)}_{sparse}]  + \mathrm{const},
    \end{aligned}
\end{equation}
where $\ell^{(i)}_{*}$ refers to the corresponding loss computed for the $i$-th sample, as illustrated in Section \ref{subsection: objective function z1}.

\subsubsection{Architecture}
Following the framework illustrated in Fig.~\ref{fig:summary} (c), we develop a modularized neural network for InDeed, as Fig.~\ref{fig: network structure} (b) shows.
According to Eq.~\eqref{eq: variational assumption}, we infer the variational posteriors of $A$ and $B$ in parallel to obtain $L$, followed by the estimation of the posterior of $S$.
The noise component ($N$) can be computed directly based on the image decomposition.
Hence, InDeed consists of two major modules, \textit{i.e.}, $f_{\theta_{L}}$ and $f_{\theta_{S}}$, respectively for the low-rank and sparsity representation, and $f_{\theta_{L}}$ is composed of two sub-modules. 
%\textcolor{red}{Following \cite{gao2022RONet}, we in practice employ ResNet \cite{he2016resnet} with group normalization as the backbone of each (sub-)module.}
%\textcolor{red}{Please refer to supplementary material for more details.} 

\textbf{For the low-rank component}, we design the low-rank module $f_{\theta_L}$ with two parallel sub-modules $f_{\theta_A} \ \mathrm{and} \ f_{\theta_B}$, following the modeling of $L=AB^T$ in Eq.~\eqref{eq: expression of L}. 
These sub-modules aim to estimate distribution parameters w.r.t. $q(A)$ and $q(B)$, namely $\phi_A=\{\mu_A,\sigma_A\} := f_{\theta_A}(Y)$ and $\phi_B=\{\mu_B,\sigma_B\} := f_{\theta_B}(Y)$.
Specifically, to implement the low-rank expression in Eq.~\eqref{eq: expression of L}, $f_{\theta_A}$ is designed to learn the mapping from the image space to the Cartesian product of $2r^0$ vector spaces, $\mathbb{R}^{h \times w} \to \prod_{i=1}^{2r^0}\mathbb{R}^{h\times 1} $, and outputs two sets of $r^{0}$ column vectors, for $\mu_A$ and $\sigma_A$, respectively. 
Similarly, $f_{\theta_B}: \mathbb{R}^{h \times w} \to \prod_{i=1}^{2r^0}\mathbb{R}^{1\times w}$ outputs two sets of $r^{0}$ row vectors for $\mu_B^T$ and $\sigma_B^T$. Then, following the reparameterization technique \cite{doersch2016vae}, we sample $\hat{A}, \hat{B}$ respectively from the inferred variational distributions, \textit{i.e.}, $q(A|\mu_A,\sigma^2_A)$ and $q(B|\mu_B,\sigma^2_B)$. 

\textbf{For the sparsity component}, we design a sparsity module $f_{\theta_S}: \mathbb{R}^{h \times w} \to \mathbb{R}^{h \times w}$ to estimate the variational parameters for $q(S|A,B)$, namely this module takes $(Y-\hat{A}\hat{B}^T)$ as inputs and outputs the corresponding parameters, \textit{i.e.}, $\phi_S=\{\mu_S,\sigma_S\} := f_{\theta_S}(Y-\hat{A}\hat{B}^T)$.

\textbf{For the noise component}, it is explicitly expressed as $\hat{N}=Y-\hat{A}\hat{B}^T -\hat{S}$, where $\hat{S}$ is sampled from $q(S|\mu_S,\sigma^2_S)$.

\textbf{For variables in $\mathcal{Z}_l$}, they are explicitly computed with Eq.~\eqref{eq: gamma formula}, \eqref{eq: omega formula}, and \eqref{eq: lambda formula}, as Fig.~\ref{fig: network structure} (b) illustrates.

For the aforementioned (sub-)modules, we in practice employ ResNet \cite{he2016resnet} with group normalization as the backbone. Please refer to Section 4 in Supplementary Material for more details of the implementation.

\subsubsection{Interpretation}
As InDeed follows the three-step framework, its interpretability comes from two key aspects: the modularized structure guided by the PGM and the additively separable loss function derived from variational inference.

First, InDeed’s structure is partially transparent, combining closed-form solutions for leaf-level variables with modularized networks ($f_{\theta}$) for middle-level variable estimation. The intermediate outputs of $f_{\theta}$, such as the distribution parameters of variational posteriors $q(A)$, $q(B)$, and $q(S)$, are statistically interpretable, given their priors and $\mathcal{Z}_l$.

Second, the loss function $\hat{\mathcal{L}}_{\mathrm{dnn}}$, derived from variational inference, integrates various terms mathematically, eliminating the need for manual balancing. Each term in $\hat{\mathcal{L}}_{\mathrm{dnn}}$ reflects the information of the intermediate outputs from different aspects, enhancing the interpretability of each module.

Finally, thanks to the separability of $\hat{\mathcal{L}}_{\mathrm{dnn}}$, we can achieve active generalization by adaptation, as Section~\ref{Section: active generalization} will illustrate.

\subsection{Error bound and active generalization} 
\label{Method:generalization}

In this section, we analyze the generalization error bound of InDeed, based on which we further develop an unsupervised test-time adaptation algorithm, for out-of-distribution (OOD) scenarios.

\subsubsection{Generalization error bound}
Let $p(Y)$ be the data distribution,  $\mathcal{D}=\{Y^{(i)}\}_{i=1}^{\vert \mathcal{D}\vert}$ be a data set sampled from $p(Y)$, 
and the  empirical distribution function defined on $\mathcal{D}$ be $\hat{p}_{\mathcal{D}}(Y)=\frac{1}{|\mathcal{D}|} \sum^{|\mathcal{D}|}_{i=1}\mathbbm{1}_{\{Y=Y^{(i)}\}}$. 
Let $\mathcal{R}(f_\theta)$  denote the expected generalization error of $f_{\theta}$ over $p(Y)$, defined as $\mathcal{R}(f_\theta) =\mathbb{E}_{p(Y)}\{\mathbb{E}_{q(\mathcal{Z}|Y)}[-\log p(Y|\mathcal{Z})]\}$,
and the empirical error over $\mathcal{D}$ as 
$\hat{\mathcal{R}}(f_\theta ; \mathcal{D})=\frac{1}{|\mathcal{D}|}\sum_{i=1}^{|\mathcal{D}|}\{\mathbb{E}_{q(\mathcal{Z}|Y^{(i)})}[-\log p(Y^{(i)}|\mathcal{Z})]\}$.
Similarly, we define empirical KL over $\mathcal{D}$ as 
$\hat{\mathrm{KL}}(q,p; \mathcal{D})=\frac{1}{|\mathcal{D}|}\sum_{i=1}^{|\mathcal{D}|}\
[\mathrm{KL}(q(\mathcal{Z}|Y^{(i)})\|p(\mathcal{Z}))]$.
Then, based on the PAC-Bayesian theorem \cite{mcallester2003pac_first, mbacke2024pac-vae, germain2016pac_Baye}, we can obtain an upper bound for $\mathcal{R}(f_\theta)$ provided in  Theorem 1.

\begin{theorem} \label{Theorem 1} 
Let $f_{\theta}$ be any $K_{\theta}$-Lipschitz-continuous function, and $\delta \in (0,1)$. We then have the following inequality hold with probability at least $1-\delta$: 
\begin{equation} \label{eq:theorem}  \small
\begin{aligned}
    \mathcal{R}(f_{\theta})    \le 
    &\ \hat{\mathcal{R}}(f_{\theta};\mathcal{D}) +  \hat{\mathrm{KL}}(q,p; \mathcal{D}) \\ & +  K \cdot d(\hat{p}_{\mathcal{D}}(Y),p(Y))+ C.
\end{aligned}
\end{equation}
Here, $d(p_1,p_2)$ refers to the discrepancy between two distributions $p_1$ and $p_2$, 
% namely, $ C_1 =\mathbb{E}_{\hat{p}_{\mathcal{D}}(Y')}\mathbb{E}_{p(Y)}d(Y,Y^{'})$ 
as defined in \cite{mbacke2024pac-vae}; $C$ is a constant given prior $p(\mathcal{Z})$;
$K=c(K_\theta)$ is determined by $K_{\theta}$.
\end{theorem}

% In Eq.~\eqref{eq:theorem}, we assume $K$ be relatively small and when $p(\mathcal{Z})$ is given, the upper bound can then be mainly controlled by the other two terms, \textit{i.e.}, $\hat{\mathcal{R}}(f_{\theta};\mathcal{D})$ and $\hat{\mathrm{KL}}(q,p; \mathcal{D})$.

To tighten the error bound, two primary approaches can be considered: reducing the Lipschitz constant $K_\theta$ \cite{fazlyab2019efficient}, and controlling the first two terms, \textit{i.e.}, $\hat{\mathcal{R}}(f_{\theta};\mathcal{D})$ and $\hat{\mathrm{KL}}(q,p; \mathcal{D})$. Herein, we mainly focus on the latter one.

\textbf{Relation to $\hat{\mathcal{L}}_{\mathrm{dnn}}$ of $f_{\theta}$}:  
By comparing Eq.~\eqref{eq: empirical loss of ELBO} and the above-mentioned upper bound, one can see that $\hat{\mathcal{R}}(f_{\theta};\mathcal{D}) +  \hat{\mathrm{KL}}(q,p; \mathcal{D})$ is the empirical version of $ \mathcal{L}_{\mathrm{dnn}}$ over $\mathcal{D}$. 
Hence, one can rewrite this upper  bound using the same notation, 
\begin{equation} 
\hat{\mathcal{L}}_{\mathrm{dnn}} (f_{\theta};\mathcal{D}) = \hat{\mathcal{R}}(f_{\theta};\mathcal{D}) +  \hat{\mathrm{KL}}(q,p; \mathcal{D}). 
\end{equation}
This loss ($\hat{\mathcal{L}}_{\mathrm{dnn}} (f_{\theta};\mathcal{D})$) is equivalent to that ($\hat{\mathcal{L}}_{\mathrm{dnn}}$) in Eq~\eqref{eq: deep loss function} excluding the supervision term ($\ell_{sup}$), in the absence of target $U$.
% When $\mathcal{D}=\mathcal{D}_{tr}$, this loss ($\hat{\mathcal{L}}_{\mathrm{dnn}} (f_{\theta};\mathcal{D})$) becomes equivalent to that ($\hat{\mathcal{L}}_{\mathrm{dnn}}$) in Eq~\eqref{eq: deep loss function} excluding the supervision term ($\ell_{\mathrm{sup}}$), which is not accessible in test time. 
Hence, minimizing Eq~\eqref{eq: deep loss function} tends to reduce the error bound in Eq.~\eqref{eq:theorem}.
\emph{This explains the good generalizability of InDeed \cite{germain2016pac_Baye} and motivates us to develop a test-time adaptation algorithm for OOD scenarios.}

\subsubsection{Active generalization for OOD scenarios}\label{Section: active generalization}
We refer to this algorithm as InDeed Active Generalization (InDeedAG). 
Furthermore, we extend InDeedAG to InDeed Online Active Generalization (InDeedOAG) to accommodate real-world scenarios necessitating adaptation for an individual OOD image only.

Let $f_{\theta^*}$ be the model trained with $\mathcal{D}_{tr}$;
let $\mathcal{D}_{ood}$ be an OOD \textit{test} data set sampled from $p(Y)$. 
When the empirical distribution over $\mathcal{D}_{ood} $ differs significantly from that of $\mathcal{D}_{tr}$, %($\hat{P}_{\mathcal{D}_{tr}}$), 
the solution of $f_{\theta^*}$ from minimization of $\hat{\mathcal{L}}_{\mathrm{dnn}} (f_{\theta};\mathcal{D}_{tr})$ can be inferior for $\mathcal{D}_{ood}$.
To improve the solution, one can intuitively further improve the model by combining the two datasets, following Theorem 1,
\begin{equation}  
f_{\theta^{\mathrm{AG}}}=\arg\min_{f_{\theta}} \hat{\mathcal{L}}_{\mathrm{AG}}:=
\hat{\mathcal{L}}_{\mathrm{dnn}} (f_{\theta};\mathcal{D}_{tr} \cup \mathcal{D}_{ood}).
\end{equation} 

Generally, reusing $\mathcal{D}_{tr}$ is inefficient or inaccessible at the test stage, thus, we adopt the loss terms in Eq~\eqref{eq: deep loss function}, \textit{i.e.}, $\ell^{(i)}_{fid},\ \ell^{(i)}_{rank}$ and $\ell^{(i)}_{sparse}$, calculated solely on $\mathcal{D}_{ood}$ for  $\hat{\mathcal{L}}_{\mathrm{AG}}$.
% From \textcolor{red}{the relationship between the upper bound} in Eq.~\eqref{eq:theorem} and the loss in Eq~\eqref{eq: deep loss function}, we adopt the loss terms in Eq~\eqref{eq: deep loss function}, \textit{i.e.}, $\ell^{(i)}_{fid},\ \ell^{(i)}_{rank}$ and $\ell^{(i)}_{sparse}$, for  $\hat{\mathcal{L}}_{\mathrm{AG}}$. 
In addition, one can solely fine-tune selected module(s), such as the sparse module ($f_{\theta_S}$) for sparsity-sensitive applications, while keeping other modules frozen, thanks to the modularized architecture of $f_\theta$ and the separability of the loss function. 
For such commonly seen scenario, we have the loss function,
\begin{equation} \label{eq:active generalization loss}
\hat{\mathcal{L}}_\mathrm{AG}:=\hat{\mathcal{L}}_\mathrm{AG_S} = \frac{1}{|\mathcal{D}_{ood}|}\textstyle\sum_{i=1}^{|\mathcal{D}_{ood}|}(  \ell_{fid}^{(i)} + \ell_{sparse}^{(i)}).
\end{equation}
Similarly, there is a less common scenario when one needs only adaptation on the low-rank module. The loss becomes,
\begin{equation}  \label{eq:active generalization loss:L}
\hat{\mathcal{L}}_\mathrm{AG}:=\hat{\mathcal{L}}_\mathrm{AG_L} = \frac{1}{|\mathcal{D}_{ood}|}\textstyle\sum_{i=1}^{|\mathcal{D}_{ood}|}(\ell_{fid}^{(i)} +\ell_{rank}^{(i)} ).
\end{equation}
Finally, one may perform the adaptation on both of the two modules, resulting in a  loss, as follows,
\begin{equation} \label{eq:active generalization loss:LS} 
\hat{\mathcal{L}}_{\mathrm{AG}}:=    
    \hat{\mathcal{L}}_{\mathrm{AG_{LS}}} = 
    \frac{1}{|\mathcal{D}_{ood}|}
   \sum_{i=1}^{|\mathcal{D}_{ood}|}(\ell^{(i)}_{fid} + \ell^{(i)}_{rank} + \ell^{(i)}_{sparse}).
\end{equation} 
Note that simultaneously fine-tuning these two modules can be challenging due to their entanglement. Section~\ref{Section4: Active generalization study} provides more discussion. 

\textbf{Online active generalization:} 
Given an OOD image during the test phase, InDeedOAG executes two sequential operations on the image, \textit{i.e.}, model adaptation through active generalization followed by inference.
InDeedOAG presents high efficiency and effectiveness, as Section~\ref{Section4: Active generalization study} will demonstrate.

To prevent model collapse and large fluctuations in $K$, we apply early-stopping \cite{blanchard2018early-stop}. Algorithms in Section 3 of Supplementary Material provide pseudocodes.

\subsection{Applications} \label{Section3: application}
We exemplify InDeed using two downstream tasks, namely, image denoising (DEN) and unsupervised anomaly detection (UAD), with specific supervision loss $\ell_{sup}$. 
Furthermore, to ensure the two low-rank components capture diverse information, we apply an additional orthogonal constraint, \begin{equation}
\ell_{orth}= \|\hat{A}^T\hat{A} - I\|^2_F + \|\hat{B}^T\hat{B} - I\|^2_F,
\end{equation}
where $\hat{A}$ and $\hat{B}$ are respectively the sampled left and right factors of low-rank components.
The overall loss function  becomes,
\begin{equation} \label{eq: network loss function}
   \hat{\mathcal{L}}_{\mathrm{DEN/UAD}} =  \hat{\mathcal{L}}_{\mathrm{dnn}}  + \tau \textstyle\sum_{i=1}^{|\mathcal{D}_{tr}|}\ell_{orth}^{(i)},
\end{equation}
where $\tau$ is a balancing weight. 
In the following, we specify the supervision loss $\ell_{sup}$ in Eq.~\eqref{eq: deep loss function} for the two tasks. 
 
\textbf{For image denoising}, we consider the target ($U \in \mathbb{R}^{h \times w}$) to be the clean image of the observation. For convenience, we use the low-rank variable $L$ instead of $A \  \mathrm{and}\  B$ for illustration.
Specifically, given $\mathcal{Z}_m=\{L,S\}$, 
the likelihood w.r.t. $U$ can be modeled as i.i.d Gaussian distributions with a hyper-parameter $\sigma_0$, namely,
$p(U|\mathcal{Z}_m) = \prod^{h,w}_{i=i, j=1} \mathcal{N}(u_{ij}|l_{ij}+s_{ij}, \sigma_0^{-1})$. 
Hence, by adopting the same reparameterization technique in Eq.~\eqref{eq:l_fid}, the supervision loss $\ell_{sup}$ can be intuitively expressed as, $\ell_{sup} = -\mathbb{E}_{q(\mathcal{Z}_m)}[\log p(U|\mathcal{Z}_m)] \approx \frac{\sigma_0}{2}\|\hat{L}+\hat{S} - U\|_F^2$.
% \begin{equation}
%     \begin{aligned}
%     \ell_{\mathrm{sup}} &= -\mathbb{E}_{q(\mathcal{Z}_m)}[\log p(U|\mathcal{Z}_m)] \\
%     &\approx \frac{\sigma_0}{2}\|\hat{L}+\hat{S} - U\|_F^2.
%     \end{aligned}
% \end{equation}}
However, minimizing this term could result in a naive solution of $\hat{L}=0$ and $\hat{S}=U$, which deviates from our assumption that $\hat{L}$ is a low-rank approximation of $U$. 
Therefore, we propose to include an additional loss of $\|\hat{L}-U\|^2_F$  to avoid such problem, namely,  
\begin{equation} \label{l_sup_DEN}
    \ell_{sup|_{\mathrm{DEN}}} = \frac{\sigma_0}{2}(\|\hat{L}+\hat{S} - U\|_F^2 + \|\hat{L}-U\|^2_F).
\end{equation}

\textbf{For unsupervised anomaly detection}, we generate abnormal samples $Y$ from normal observations $U_N \in \mathbb{R}^{h \times w}$ and synthetic anomalies $U_A \in \mathbb{R}^{h \times w}$, by a self-supervised scheme (\textit{Please refer to Section 4.3 of Supplementary Material}). Specifically, $U_A$ is expressed as $U_A = Y - U_N$ to satisfy the decomposition. 
Thereby, we have the ground truth $U = \{U_N, U_A\}$, 
and $p(U_A,U_N|\mathcal{Z}_m)=p(U_A|S)p(U_N|L)$. 
We model $U_A$ as i.i.d Gaussian distributions with a user-determined hyper-parameter $\sigma_0$, namely, $p(U_A|S)= \prod^{h,w}_{i=1, j=1} \mathcal{N}(U_{a_{ij}}|s_{ij}, \sigma_0^{-1})$.
Similarly, we define $p(U_N|L)= \prod^{h,w}_{i=1, j=1} \mathcal{N}(U_{n_{ij}}|l_{ij}, \sigma_0^{-1})$. Hence, the supervision loss for UAD is given by, 
\begin{equation} \label{loss: AD supervision term}
    \ell_{sup|_{\mathrm{UAD}}} = \frac{\sigma_0}{2}\left(\|\hat{L} - U_N\|_F^2 + \|\hat{S} - U_A\|_F^2 \right).  
\end{equation}

\begin{table*}
\centering
\caption{Ablation Study of InDeed with different backbones, depth-kernel pairs  ($d, ks$), maximal rank $r^0$, normalization strategies, and loss functions.
The results of $L$ present metrics calculated via $L$ and $U$, indicating low-rank estimation quality. $L+S$ refers to the recovery image, with its results as denoising performance.
\textit{BatchNorm: Batch Normalization; GroupNorm: Group Normalization}.
Bold fonts indicate the best results.} \label{Table:ablation study}
\begin{tabular}{|c|lllll|ll|ll|l|}
\hline
\multicolumn{1}{|c|}{\multirow{2}{*}{Model}} & \multicolumn{5}{c|}{Settings}                                                                                                                                                                                   & \multicolumn{2}{c|}{$L$}           & \multicolumn{2}{c|}{$L+S$} & \multicolumn{1}{c|}{\multirow{2}{*}{\#Para}} \\ \cline{2-10}
\multicolumn{1}{|c|}{}                       & \multicolumn{1}{l|}{Backbone}                & \multicolumn{1}{l|}{$(d,ks)$}                   & \multicolumn{1}{l|}{$r^0$} & \multicolumn{1}{l|}{Norm}                       & loss function      & PSNR $\uparrow$          & SSIM $\uparrow$           & PSNR $\uparrow$      & SSIM $\uparrow$       & \multicolumn{1}{c|}{}                        \\ \hline
\#1                                          & \multicolumn{1}{l|}{\multirow{3}{*}{ResNet}} & \multicolumn{1}{l|}{(15, 7)}                   & \multicolumn{1}{l|}{\multirow{3}{*}{64}}  & \multicolumn{1}{l|}{\multirow{3}{*}{GroupNorm}} & \multirow{3}{*}{$\hat{\mathcal{L}}_{\mathrm{DEN}}$ in Eq.~\eqref{eq: network loss function}} & 22.27          & 0.6335          & 31.69      & 0.9217      & 8.8M                                         \\
\#2                                          & \multicolumn{1}{l|}{}                        & \multicolumn{1}{l|}{(21, 5)}                  & \multicolumn{1}{l|}{}                     & \multicolumn{1}{l|}{}                           &                    & 22.67          & 0.6612          & 28.54      & 0.8744      & 7.5M                                         \\
\textbf{\#3}                                          & \multicolumn{1}{l|}{}                        & \multicolumn{1}{l|}{(35, 3)}                  & \multicolumn{1}{l|}{}                     & \multicolumn{1}{l|}{}                           &                    & \textbf{22.69} & \textbf{0.6618} & \textbf{32.32}      & \textbf{0.9278}      & 5.8M                                         \\ \hline
\#4                                          & \multicolumn{1}{l|}{\multirow{5}{*}{ResNet}} & \multicolumn{1}{l|}{\multirow{5}{*}{(35, 3)}} & \multicolumn{1}{l|}{4}                    & \multicolumn{1}{l|}{\multirow{5}{*}{GroupNorm}} & \multirow{5}{*}{$\hat{\mathcal{L}}_{\mathrm{DEN}}$ in Eq.~\eqref{eq: network loss function}} & 17.08          & 0.4803          & 31.59      & 0.8768      & 5.8M                                         \\
\#5                                          & \multicolumn{1}{l|}{}                        & \multicolumn{1}{l|}{}                         & \multicolumn{1}{l|}{8}                    & \multicolumn{1}{l|}{}                           &                    & 18.38          & 0.5065          & 31.58      & 0.8769      & 5.8M                                         \\
\#6                                          & \multicolumn{1}{l|}{}                        & \multicolumn{1}{l|}{}                         & \multicolumn{1}{l|}{16}                   & \multicolumn{1}{l|}{}                           &                    & 19.50          & 0.5088          & 31.70      & 0.8786      & 5.8M                                         \\
\#7                                          & \multicolumn{1}{l|}{}                        & \multicolumn{1}{l|}{}                         & \multicolumn{1}{l|}{32}                   & \multicolumn{1}{l|}{}                           &                    & 21.12          & 0.5610          & 31.71      & 0.8790      & 5.8M                                         \\
\#8                                          & \multicolumn{1}{l|}{}                        & \multicolumn{1}{l|}{}                         & \multicolumn{1}{l|}{96}                   & \multicolumn{1}{l|}{}                           &                    & 20.96          & 0.5625          & 31.58      & 0.8792      & 5.8M                                         \\ \hline
\#9                                          & \multicolumn{1}{l|}{ResNet}                  & \multicolumn{1}{l|}{(35, 3)}                  & \multicolumn{1}{l|}{64}                   & \multicolumn{1}{l|}{GroupNorm}                  & $\hat{\mathcal{L}}_{\mathrm{DEN}}$ w/o $\ell_{rank}, \ell_{sparse},\ell_{fid}$         & 13.01          & 0.5180          & 32.16      & 0.8891      & 5.8M                                         \\ \hline
\#10                                         & \multicolumn{1}{l|}{ResNet}                  & \multicolumn{1}{l|}{(35, 3)}                  & \multicolumn{1}{l|}{64}                   & \multicolumn{1}{l|}{BatchNorm}                  & $\hat{\mathcal{L}}_{\mathrm{DEN}}$ in Eq.~\eqref{eq: network loss function}                  & 12.18          & 0.4030          & 31.91      & 0.8819      & 5.8M                                         \\ \hline
\#11                                         & \multicolumn{1}{l|}{Transformer}             & \multicolumn{1}{l|}{N/A}                      & \multicolumn{1}{l|}{64}                   & \multicolumn{1}{l|}{N/A}                        & $\hat{\mathcal{L}}_{\mathrm{DEN}}$ in Eq.~\eqref{eq: network loss function}                  & 21.00          & 0.5570          & 30.87      & 0.8713      & 79M                                          \\ \hline
\end{tabular}
\end{table*}

\section{Experiments}
In this section,  we first introduced data and implementation details in Section~\ref{Section 4.1} and performed ablation studies in Section~\ref{Section4: Ablation study}. Then, we evaluated the performance of InDeed on two tasks, \textit{i.e.}, image denoising and unsupervised anomaly detection (UAD) in Section~\ref{Section4:image denoising} and~\ref{Section4:anomaly detection}, respectively. 
Subsequently, in Section~\ref{Section4: Active generalization study} we further studied the performance of active generalization in out-of-distribution scenarios. Finally, in Section~\ref{Section4:interpretability} we discussed the interpretability of each module and analyzed their contribution to model performance. \emph{For more results and visualization, please refer to Section 5 in Supplementary Material.}

\subsection{Data and implementation details}
\label{Section 4.1} 
\textbf{For image denoising}, we utilized the DIV2K dataset \cite{Agustsson_2017_CVPR_Workshops} for model training and ablation studies, consisting of 800 training and 100 validation high-resolution images. Gaussian white noise (AGWN) with noise levels ranging between $[0,75]$ was added to generate noisy images. For in-distribution denoising evaluation, we used three widely-used high-quality datasets: CBSD68 \cite{DEN-dataset-BSD}, Kodak24 \cite{DEN-dataset-Kodak24}, and McMaster \cite{DEN-dataset-McM}. To assess generalizability, we employed two real-world noisy datasets: SIDD (small version) \cite{SIDD_2018_CVPR} with 160 images from 10 scenes using five smartphone cameras, and PolyU \cite{xu2018polyU}, containing 100 images from 40 scenes using five cameras. Two metrics were used to evaluate the performance, \textit{i.e.}, the standard Peak Signal to Noise Ratio (PSNR) and the Structural Similarity (SSIM) index.

\textbf{For unsupervised anomaly detection}, we used MVTecAD \cite{bergmann2019mvtec} for training and in-distribution evaluation. MVTecAD includes 15 object classes with 3629 training and 1725 test images. Each class is further divided based on the type of anomaly, resulting in a total of 73 defect subclasses (excluding the "good" subclass). To assess generalizability, we used three OOD datasets: noisy MVTecAD, Severstal steel images \cite{severstal2019steel}, and the Medical OOD (MOOD) dataset \cite{cao2020benchmarkmood}. Noisy MVTecAD was generated by adding AGWN ($\sigma = 2.55$) to MVTecAD test images. The Severstal dataset is a steel defect detection dataset, and we randomly selected 256 images for evaluation. Finally, the MOOD dataset is for OOD medical image analysis. We evaluated the models on its validation set with 5 brain MRI toy cases. We adopt two commonly used metrics \cite{zavrtanik_draem_2021} to evaluate the performance, \textit{i.e.}, the average precision (AP), and the area under ROC curve (AUROC) scores. 
\emph{Note that when the positive and negative data are severely imbalanced, the AUROC metric may be inflated \cite{davis2006relationship}, while the AP score, which summarizes the precision-recall (PR) curve, can indicate the level of false positives.}
% AUROC scores can sometimes be inflated due to the imbalance between positive and negative data in UDA \cite{davis2006relationship}, while the AP score serves as a crucial metric that summarizes the precision-recall (PR) curve, reflecting the rate of false positives. Therefore, both metrics are important.

In the training stage, we set the image size $h\times w$ as $128 \times 128$.
For the hyper-parameters, we set $\alpha^0_{\gamma}=\alpha^0_{\omega}=\alpha^0_{\lambda}=2$,  $\beta^0_{\gamma}= \beta^0_{\omega}=1 \times e^{-6}$ and $\beta^0_{\lambda}=1 \times e^{-8}$ in Eq.~\eqref{eq:model gamma}, \eqref{eq:model omega} and \eqref{eq:model lambda}, $r^0=64$ in Eq.~\eqref{eq: expression of L}, and the balancing parameters $\tau=1$ in Eq.~\eqref{eq: network loss function}. 
We trained our models using 
the ADAM \cite{kingma2014adam} optimizer for 2000 epochs.
The initial learning rate was $1 \times 10^{-4}$, and decayed by a factor of 0.5 every 200 epochs.
During testing, we divided each image into overlapping $128 \times 128$ patches if not divisible by 128, and stitched the results to form the final prediction at the original size.
The proposed method was implemented via Pytorch, and all models were trained and tested on an RTX 3090 GPU with 24 GB memory.

\begin{table}
\centering
\caption{In-distribution study of color image denoising with different noise level ($\sigma$). Bold fonts indicate the best results and the italics indicate the second-best result.}
\label{Table-DEN-Color}
\resizebox{0.95\linewidth}{!}{
\begin{tabular}{l@{ }|l@{ }|l@{ }l|l@{ }l|l@{ }l|l@{ }l}
\hline
\multicolumn{1}{c|}{\multirow{2}{*}{Dataset}} & \multicolumn{1}{c|}{\multirow{2}{*}{Method}} & \multicolumn{2}{c|}{$\sigma = 15$}          & \multicolumn{2}{c|}{$\sigma = 25$}          & \multicolumn{2}{c|}{$\sigma = 35$}          & \multicolumn{2}{c}{$\sigma = 50$}           \\ \cline{3-10} 
\multicolumn{1}{c|}{}                         & \multicolumn{1}{c|}{}                        & PSNR           & SSIM            & PSNR           & SSIM            & PSNR           & SSIM            & PSNR           & SSIM            \\ \hline
CBSD68                                        & CBM3D \cite{DEN-BM3D}                                       & 33.52          & .9248          & 30.71          & .8716          & 28.89          & .8207          & 27.38          & .7669          \\
& CDnCNN  \cite{DEN-DnCNN}                                     & 33.89          & .9317          & 31.23          & .8863          & 29.58          & .8452          & 27.92          & .7915          \\
& FFDNet \cite{DEN-FFDNet}                                      & 33.87          & .9318          & 31.21          & .8857          & 29.58          & .8445          & 27.96          & .7916          \\
& RONet-C  \cite{gao2022RONet}                                    & 33.99          & \textit{.9336} & 31.36 & .8902          & 29.74 & \textit{.8514} & 28.14          & .8009          \\
&VIRNet \cite{VIRNET} & \textbf{34.27} & \textbf{.9340} & \textit{31.65}          & \textit{.8918}          & \textbf{30.04} & \textbf{.8548} & \textbf{28.45} & \textbf{.8083}  \\
&InDeed   & \textit{34.06}          & .9329          & \textbf{31.67} & \textbf{.8990} & \textit{29.98}          & .8483          & \textit{28.15}          & \textit{.8020}           \\ \hline
Kodak24                                       & CBM3D \cite{DEN-BM3D}                                       & 34.28          & .9164          & 31.68          & .8682          & 29.90          & .8212          & 28.46          & .7751          \\
 & CDnCNN  \cite{DEN-DnCNN}                                     & 34.48          & .9212          & 32.03          & .8774          & 30.46          & .8390          & 28.85          & .7895          \\
 & FFDNet  \cite{DEN-FFDNet}                                     & 34.64          & .9230          & 32.13          & .8790          & 30.57          & .8407          & 28.98          & .7929          \\
& RONet-C  \cite{gao2022RONet}                                    & 34.80 & .9254 & 32.33        & .8845          & 30.77          & \textit{.8484}          & 29.18         & .8020          \\
&VIRNet \cite{VIRNET} & \textit{35.15}          & \textbf{.9283} & \textbf{32.75} & \textit{.8908}          & \textbf{31.22} & \textbf{.8583} & \textbf{29.69} & \textbf{.8186}  \\
&InDeed   & \textbf{35.15} & \textit{.9259}          & \textit{32.66}          & \textbf{.9000} & \textit{31.16}          & .8436          & \textit{29.43}          & \textit{.8095}           \\ \hline
McMaster                                      & CBM3D  \cite{DEN-BM3D}                                      & 34.06          & .9150           & 31.66          & .8739          & 29.92          & .8327          & 28.51          & .7934          \\
 & CDnCNN  \cite{DEN-DnCNN}                                   & 33.44          & .9070           & 31.51          & .8724          & 30.14          & .8412          & 28.61          & .7985          \\
& FFDNet  \cite{DEN-FFDNet}                                     & 34.66          & .9247          & 32.35          & .8891          & 30.81          & .8573          & 29.18          & .8157          \\
& RONet-C  \cite{gao2022RONet}                                    & 34.77 & .9251 & 32.51          & .8920          & 31.00 & \textit{.8627} & 29.39 & .8245          \\
&VIRNet  \cite{VIRNET} & \textbf{35.32} & \textbf{.9312} & \textit{33.08}          & \textbf{.9016} & \textbf{31.59} & \textbf{.8758} & \textbf{30.02} & \textbf{.8434}  \\ 
&InDeed   & \textit{35.07}          & \textit{.9256}          & \textbf{33.52} & \textit{.8926}          & \textit{31.22}          & .8438         & \textit{29.50}          & \textit{.8345}       \\ \hline
\end{tabular}}
\end{table}

\begin{table*}[!ht]
\centering
\caption{Generalizability of image denoising on two out-of-distribution datasets, \textit{i.e.}, SIDD and PolyU. 
% The first two rows refer to the generalizability of the comparison SOTA method RONet and the proposed method. 
ADP refers to the adaptation time. 
Bold font indicates the best result within each subtable, and an underline denotes the best result across the entire table.
% Note that the bold results refer to the SOTA in terms of generalizability on OOD datasets while the bold and underlined results refer to the SOTA after active generalization. \textcolor{red}{TBC}
}
\label{Table: generalization for DEN}
\begin{threeparttable}
\resizebox{1\textwidth}{!}{
\begin{tabular}{c|l|llllllc|lllll|l|c}
\hline
\multirow{2}{*}{Method}  & \multicolumn{1}{c|}{\multirow{2}{*}{Metric}} & \multicolumn{7}{c|}{SIDD}                                                                                                                                                                                          & \multicolumn{7}{c}{PolyU}                                                                                                                                                                        \\ \cline{3-16} 
                         & \multicolumn{1}{c|}{}                        & G4                    & GP                    & IP                    & N6                    & \multicolumn{1}{l|}{S6}                    & \multicolumn{1}{l|}{mean}               & \multicolumn{1}{l|}{ADP} & C1                    & C2                    & C3                    & C4                    & C5                    & \multicolumn{1}{l|}{mean}               & \multicolumn{1}{l}{ADP}      \\ \hline
\multicolumn{16}{c}{(A) OOD performance without active generalization } \\ \hline
\multirow{2}{*}{\shortstack{MaskedDen$_{[\sigma=15]}$ \cite{chen2023masked}}}& PSNR                                         & 34.79$^*$  & 33.21$^*$  & 34.86$^*$  & 29.60$^*$       & \multicolumn{1}{l|}{29.25$^*$}        & \multicolumn{1}{l|}{32.56$^*$}        & \multirow{2}{*}{-}       & 32.69  & 34.15  & 35.05  & 34.98        & 32.50& \multicolumn{1}{l|}{33.80}        & \multirow{2}{*}{\textbf{-}}  \\
& SSIM   & .8777$^*$ & .8160$^*$ & .8680$^*$ & .6307$^*$      & \multicolumn{1}{l|}{.6650$^*$}       & \multicolumn{1}{l|}{.7812$^*$}       &                         & .9561 & .9675 & .9288 & .9366 & .9138       & \multicolumn{1}{l|}{.9387}       &                              \\ \hline 

\multirow{2}{*}{RONet-C \cite{gao2022RONet}}   & PSNR                                         & 30.16                 & 29.83                 & 29.40                 & 26.30                 & \multicolumn{1}{l|}{25.87}                 & \multicolumn{1}{l|}{28.36}                 & \multirow{2}{*}{-}       & 31.74                 & 35.00                 & 34.17                 & 34.19                 & 32.42                 & \multicolumn{1}{l|}{33.22}                 & \multirow{2}{*}{-}           \\
                         & SSIM                                         & .6281                & .6203                & .5702                & .4499                & \multicolumn{1}{l|}{.4621}                & \multicolumn{1}{l|}{.5461}                &                          & .9291                & .9232                & .9380                & .9246                & .8868                & \multicolumn{1}{l|}{.9198}                &                              \\ \hline
\multirow{2}{*}{VIRNet \cite{VIRNET}}   & PSNR                                         & 29.64                 & 29.64                 & 28.91                 & 25.92                 & \multicolumn{1}{l|}{25.70}                 & \multicolumn{1}{l|}{28.02}                 & \multirow{2}{*}{-}       & 34.09                 & \textbf{36.89}                 & 36.73                 & 35.95                 & 32.80                 & \multicolumn{1}{l|}{34.98}                 & \multirow{2}{*}{-}           \\
& SSIM                                         & .6179                & .6160                & .5594                & .4397                & \multicolumn{1}{l|}{.4512}                & \multicolumn{1}{l|}{.5368}                &                          & .8865                & .9146                & .9180               & .9164                & .8400                & \multicolumn{1}{l|}{.8930}                &                              \\ \hline
\multirow{2}{*}{InDeed}    & PSNR                                         & \textbf{31.42}        & \textbf{30.88}        & \textbf{30.94}        & \textbf{27.29}        & \multicolumn{1}{l|}{ \textbf{27.24}}        & \multicolumn{1}{l|}{\textbf{29.68}}        & \multirow{2}{*}{-}       & \textbf{34.67}        & 36.47& \textbf{36.74}        & \textbf{36.06}        & \textbf{33.47}        & \multicolumn{1}{l|}{\textbf{35.26}}        & \multirow{2}{*}{\textbf{-}}  \\
                         & SSIM                                         & \textbf{.6878}       & \textbf{.6564}       & \textbf{.6551}       & \textbf{.4921}       & \multicolumn{1}{l|}{\textbf{.5115}}       & \multicolumn{1}{l|}{\textbf{.6054}}       &                          & \textbf{.9363}       & \textbf{.9326}       & \textbf{.9244}       & \textbf{.9219}       & \textbf{.8920}       & \multicolumn{1}{l|}{\textbf{.9210}}       &                              \\ 
                         \hline 
\multicolumn{16}{c}{(B) OOD Performance with active generalization (AG)} \\ \hline

\multirow{2}{*}{InDeedAG} & PSNR                                         & \textbf{34.39}  & \textbf{33.11}  & \textbf{34.02}  & \textbf{29.65}  & \multicolumn{1}{l|}{ \textbf{29.55}}  & \multicolumn{1}{l|}{ \textbf{32.29}}  & \multirow{2}{*}{28s}      &  \textbf{{\ul 35.94}}  &  \textbf{{\ul 38.75}} & \textbf{38.34}  & \textbf{37.24}  & \textbf{35.30} & \multicolumn{1}{l|}{ \textbf{36.75}}  & \multirow{2}{*}{20s} \\
                         & SSIM                                         &  \textbf{.8084} &  \textbf{.7488} &  \textbf{.8016} &  \textbf{.5914} & \multicolumn{1}{l|}{ \textbf{.6103}} & \multicolumn{1}{l|}{ \textbf{.7206}} &                          & {\ul \textbf{.9613}} &  \textbf{.9567} &  \textbf{.9547} &  \textbf{.9544} &  \textbf{.9324} & \multicolumn{1}{l|}{ {\ul \textbf{.9521}}} &   \\[0.3ex]  \hline

\multirow{2}{*}{InDeedAG$\beta$ (L)} & PSNR                                         & 31.13                 & 30.20                 & 31.06                 & 27.49                 & \multicolumn{1}{l|}{27.67}                 & \multicolumn{1}{l|}{29.67}                 & \multirow{2}{*}{21s}      & 35.27                 & 37.27                 & 37.37                 & 36.73                 & 34.14                 & \multicolumn{1}{l|}{35.93}                 & \multirow{2}{*}{17s}          \\
                         & SSIM                                         & .7584                & .6808                & .7156                & .5572                & \multicolumn{1}{l|}{.5733}                & \multicolumn{1}{l|}{.6598}                &                          & .9396                & .9314                & .9371                & .9365                & .9014                & \multicolumn{1}{l|}{.9300}                &                              \\ \hline
\multirow{2}{*}{InDeedAG$\beta$ (LS)} & PSNR                                         & {31.83}  & {31.14}  & {31.17}  & {27.56}  & \multicolumn{1}{l|}{27.54}  & \multicolumn{1}{l|}{29.94}  & \multirow{2}{*}{22s}      & {35.31}  & {37.46}  & {37.32}  & {36.63}  & {34.18}  & \multicolumn{1}{l|}{35.92}  & \multirow{2}{*}{13s} \\
                         & SSIM                                         & {.7032} & {.6656} & {.6644} & {.5009} & \multicolumn{1}{l|}{.5220} & \multicolumn{1}{l|}{.6149} &                          & {.9437} & {.9390} & {.9387} & {.9409} & {.9045} & \multicolumn{1}{l|}{{.9340}} &   \\ \hline

\multicolumn{16}{c}{(C) OOD Performance with online active generalization (OAG)} \\ \hline
\multirow{2}{*}{InDeedOAG} & PSNR & {\ul \textbf{35.94}}      & {\ul \textbf{34.63}}      & {\ul \textbf{36.03}}      & {\ul \textbf{32.95}}      & \multicolumn{1}{l|}{{\ul \textbf{32.24}}}      & \multicolumn{1}{l|}{{\ul \textbf{34.48}}}      & 0.07s/ &  \textbf{35.67}  &  \textbf{38.66}      & {\ul \textbf{38.48}}      & {\ul \textbf{37.49}}        & {\ul \textbf{35.68}}        & {\ul \textbf{36.84}}    & 0.04s/ \\
& SSIM  & {\ul \textbf{.9043}}     & {\ul \textbf{.8622}}     & {\ul \textbf{.8913}}     & {\ul \textbf{.8061}}     & \multicolumn{1}{l|}{{\ul \textbf{.7931}}}     & \multicolumn{1}{l|}{{\ul \textbf{.8534}}}     &     image                      & \textbf{.9520} & {\ul \textbf{.9573}}     & {\ul \textbf{.9554}}     & {\ul \textbf{.9550}}       & {\ul \textbf{.9414}}       &  \textbf{.9517}   &   image                   \\[0.3ex]    \hline
\end{tabular}
}
\begin{tablenotes}
        \footnotesize
        \item *MaskedDen$_{[\sigma=15]}$ was trained with Gaussian noise at $\sigma=15$, similar to that of SIDD. Hence, these results in (A) are not considered as OOD.
      \end{tablenotes}
  \end{threeparttable}
\end{table*}

\subsection{Ablation study} \label{Section4: Ablation study}
We employed the image denoising task and DIV2K dataset for this study.
The training set and the validation set at the noise level of $\sigma=25$ were respectively used for our models. 
%We trained models with the DIV2K training dataset and validated them with the validation set of  DIV2K at noise level $\sigma=25$.
% As Table~\ref{Table:ablation study} illustrates, we investigated five factors, including the network structure, namely depth and kernel size $(d,ks)$, the rank-controlling hyperparameter $r^0$ in Eq.\eqref{eq:A modeling}, the proposed loss function $\hat{\mathcal{L}}_{\mathrm{DEN}}$ in Eq.~\eqref{eq: network loss function}, the types of the normalization strategy and network backbones.

We first investigated the impact of model depth $d$ and kernel size $ks$ for $f_{\theta_L}$, adjusting them jointly to maintain the full receptive field. Three configurations of $(d, ks)$ were explored, and Models \#1-3 were trained accordingly. Model \#3 achieved the best performance at $(d, ks) = (3, 35)$.
Next, we examined the effect of the maximum rank $r^0$. As $r^0$ increased, the performance of $L$ and $L+S$ peaked at $r^0 = 64$, which is close to the average rank of clean images in DIV2K, estimated as 73 via SVD. The performance of $L$ was more sensitive to changes in $r^0$, as its representation is limited by $r^0$, while $L+S$ can be compensated by $f_{\theta_S}$.

To study the impact of loss functions, we trained Model \#9 by minimizing $\hat{\mathcal{L}}_{\mathrm{DEN}}$ without $\ell_{rank}$, $\ell_{sparse}$ and $\ell_{fid}$.
While its performance on $L+S$ was comparable to Model \#3, the $L$ component deviated from the optimal low-rank representation, leading to a less interpretable end-to-end network.
To evaluate the utility of GroupNorm, Model \#10 was trained with BatchNorm instead. 
The results indicated that GroupNorm was essential for capturing an optimal low-rank representation, since BatchNorm, applying normalization over a mini-batch, was less suited for the independent modeling of column vectors in $A$ and $B$ in Eq.~\eqref{eq:A modeling} and \eqref{eq:B modeling}. 
GroupNorm, normalizing channel-wise, was then proved more appropriate here.
% The result showed that GroupNorm was essential for capturing an ideal low-rank representation, as BatchNorm, which normalizes over a mini-batch, was less suited to the independent modeling of column vectors in $A$ and $B$ in Eq.~\eqref{eq:A modeling} and Eq.~\eqref{eq:B modeling}. GroupNorm, normalizing channel-wise, proved more appropriate for our model.
Finally, we replaced the backbone with a transformer in Model \#11, which nevertheless led to an evident performance drop.

Based on the above study, we adopted the setting of Model \#3 for the following studies unless stated otherwise.

\subsection{General performance}
\subsubsection{Image denoising} \label{Section4:image denoising}
In this section, we studied the performance of InDeed on both Gaussian denoising and real-world denoising tasks. We trained InDeed with noisy DIV2K images, with the noise level $\sigma$ of AGWN uniformly distributed in $(0,75]$.

\textbf{In-distribution performance of Gaussian denoising.} 
We evaluated InDeed on CBSD68, Kodak24, and McMaster at different noise levels ($\sigma \in \{15, 25, 35, 50\}$) and compared it with five SOTA methods: BM3D \cite{DEN-BM3D}, DnCNN \cite{DEN-DnCNN}, FFDNet \cite{DEN-FFDNet}, RONet-C \cite{gao_rank-one_2022}, and VIRNet \cite{VIRNET}. 

\begin{figure}
    \centering
    \includegraphics[width=1\linewidth]{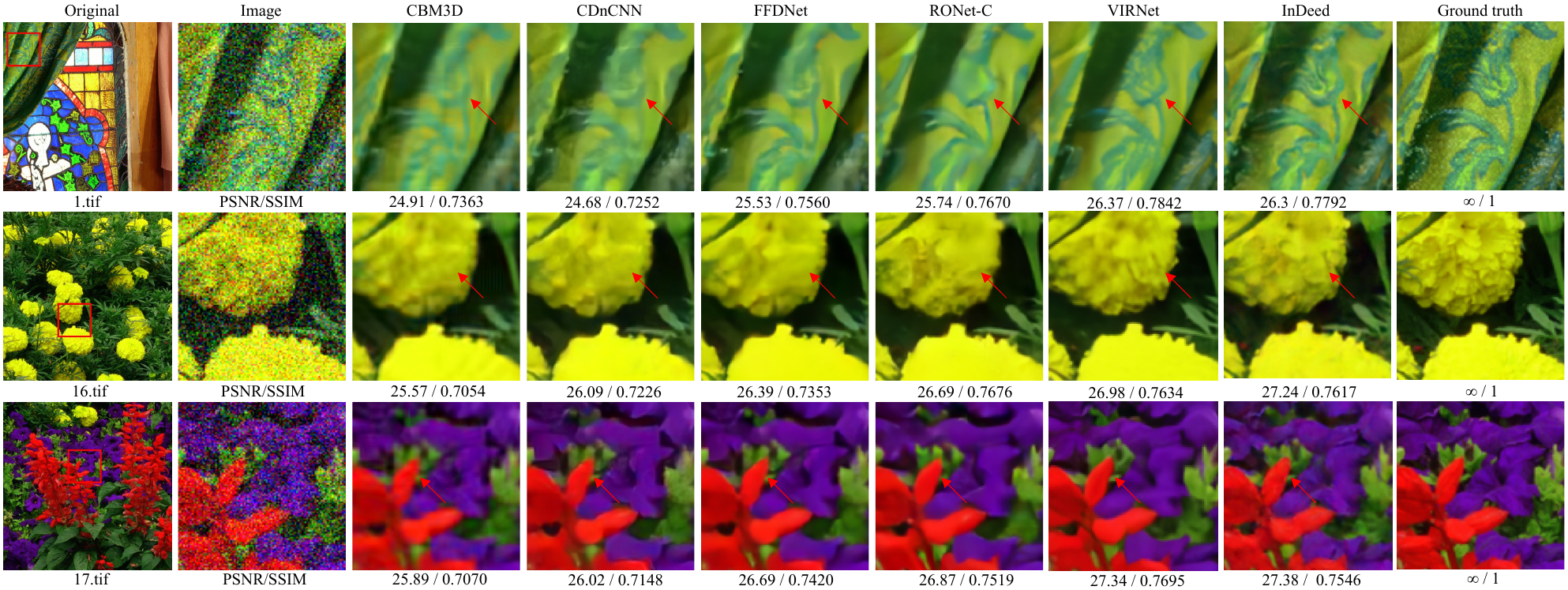}
    \caption{Visualization of color image denoising under $\sigma=50$. Three typical examples (cropped patches of size $100 \times 100$, denoted by red boxes) from McMaster. Please zoom in the online electronic version for more details.}
    \label{Figure-results-DEN_VIS_Color}
\end{figure}

Table~\ref{Table-DEN-Color} shows the results. One can see that the best performance in each task was achieved by either InDeed or VIRNet. 
One particular phenomenon occurs at $\sigma=25$, where InDeed outperforms VIRNet in four out of six metrics.
% Another point worth noting is the relatively low SSIM, particularly at a noise level of $\sigma = 35$.
% This is because the similarity between image details and noise in the input of $f_{\theta_S}$ (namely, $Y-S$) \textcolor{red}{(wfp: how can you prove this? If cannot, I prefer no mention of it.)}complicates the extraction of sparse component $S$, causing some image details to be misallocated as noise ($N=Y-L-S$), which reduces the SSIM. 
%Fig.~\ref{Figure-results-DEN_VIS_Color} presents the visualization results of three typical examples. One can see that InDeed's denoising results more closely align with the content of the clean image, offering a more authentic representation, including texture patterns and edges, as pointed out in the figures. In contrast, other models tend to produce overly smooth reconstructions, even though their PSNR/SSIM metrics could be better.
Fig.~\ref{Figure-results-DEN_VIS_Color}  provides three typical examples from our study, showing the visual difference of the denoised results by the compared methods.
Furthermore, we implemented InDeed for gray-scale Gaussian denoising, which demonstrated a comparable performance, achieving the best scores in 16 metrics (out of 24). \emph{Please refer to Section 5.1 in Supplementary Material for more details.}

\textbf{OOD generalizability in real-world denoising.} 
We evaluated InDeed on SIDD and PolyU datasets without retraining, and compared with RONet-C \cite{gao_rank-one_2022}, VIRNet \cite{VIRNET}, and MaskedDen \cite{chen2023masked}, which were particularly developed for OOD generalization.
Table~\ref{Table: generalization for DEN} (A) presents the results. 
InDeed performed better than RONet and VIRNet in 23 out of 24 metrics 
(only in C2 of PolyU, VIRNet had a better PSNR score).  
This indicates the superior generalizability of explicit modeling of image decomposition. 
Note that MaskedDen achieved the best results in the SIDD dataset, which could not be considered as an OOD scenario, 
and it experienced a significant performance drop in  PolyU when it was  OOD. % with a lower average PSNR, recording the second-worst result.
This is because MaskedDen was trained with the setting of Gaussian noise level as $\sigma=15$, which was close to the noise level of SIDD (thus considered as in-distribution). 
%Moreover, due to the feasibility of reducing the generalization error bounded in Theorem~\ref{Theorem 1}, the performance of InDeed can be efficiently improved via active generalization during the test phase, which will be demonstrated in Section~\ref{Section4: Active generalization study}. ——庄吓海：感觉在这里写这句话有点造成不必要的影响

\subsubsection{Unsupervised anomaly detection} \label{Section4:anomaly detection}
We first trained one single model with all classes from MVTecAD, following uniAD \cite{uniAD}, and then compared it with six state-of-the-art (SOTA) methods, including uniAD \cite{uniAD}, DRAEM \cite{zavrtanik_draem_2021}, FAVAE \cite{dehaene2020FAVAE}, FastFlow \cite{yu2021fastflow}, CutPaste \cite{li_cutpaste_nodate}, and STPM \cite{wang2021student} in both in-distribution and OOD scenarios. 
We utilized the official model of UniAD and retrained the other five SOTA models under the same setting.
% Since the uniAD was specifically developed for UUAD, we directly utilized its official model for testing. For fair comparisons, the other five SOTA models were retrained under the same settings as InDeed. 
Specifically, we clustered the 73 defect classes into three groups, namely, Low, Mid, and High, according to the rank of test images estimated by SVD, to better present the inherent variability of UAD in terms of image rank.

\textbf{In-distribution performance.}
Table~\ref{Table:AD} presents results on the MVTecAD test dataset. 
One can see InDeed and uniAD ranked the top two in all categories.  
When solely comparing these two methods, one can conclude that uniAD tended to produce more false-positive predictions, 
as it obtained marginally better AUROC (\textit{e.g.}, Mid: 94.8\% vs. 94.6\%) but evidently lower AP values (\textit{e.g.}, Mid: 66.1\% vs. 74.8\%), compared to InDeed.

\begin{figure}
    \centering
    \includegraphics[width=1\linewidth]{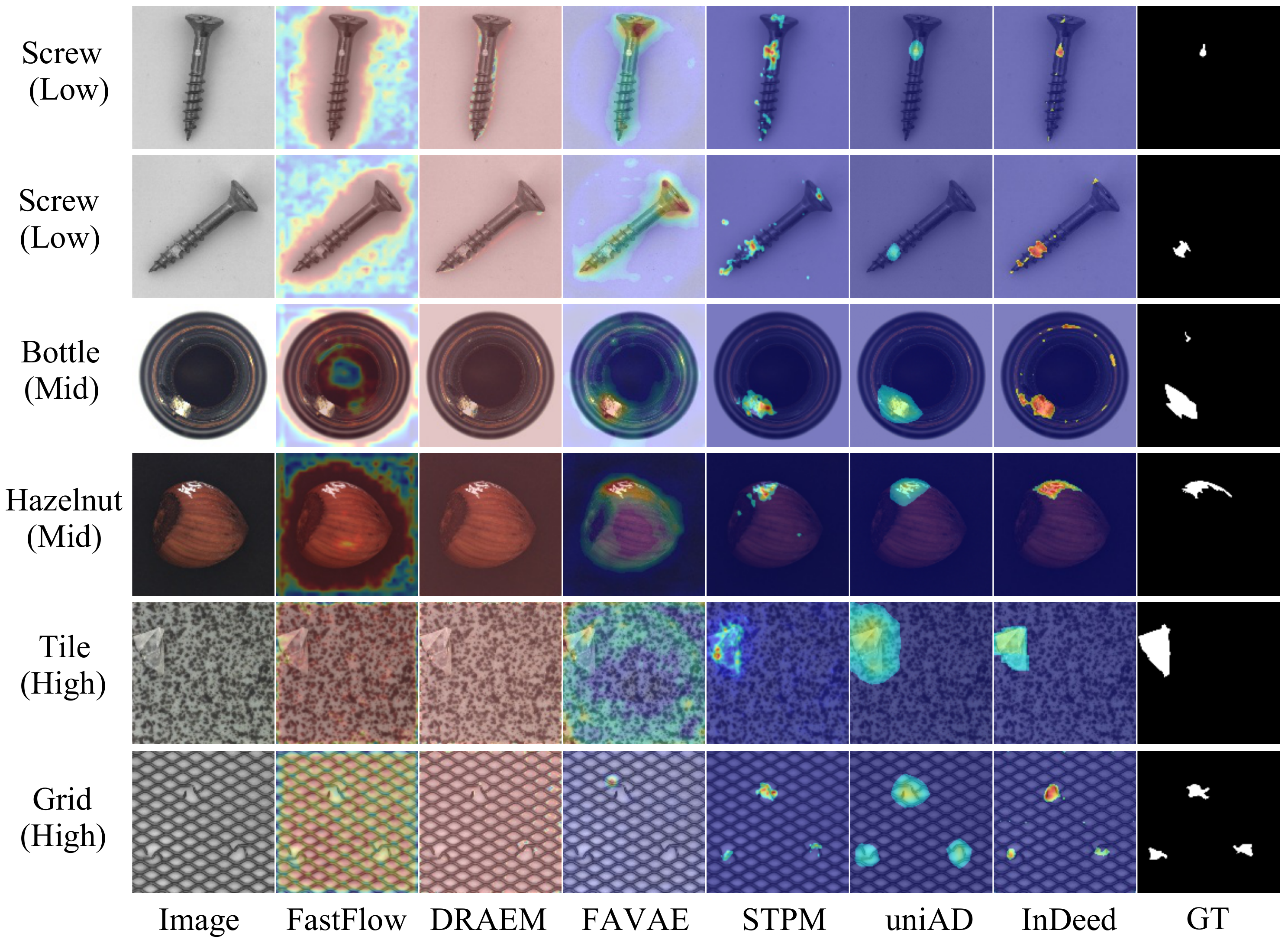}
    \captionsetup{font=small}
    \caption{Examples from MVTecAD and the anomaly map overlay. 
    \emph{Note that the color of original images may look different when they are overlaid with the anomaly maps, due to the mixture of colors}.
    GT: ground truth segmentation map. 
    }
    \label{fig:qualitative}
\end{figure}

Fig.~\ref{fig:qualitative} visualizes the results of six samples selected from the three groups by all the compared methods. 
Both InDeed and uniAD performed well in these cases, though uniAD could generate more false positives.
This is consistent with the previous conclusion from Table~\ref{Table:AD}.
One may also find that InDeed could be challenged in distinguishing these normal pixels exhibiting sparsity property from anomalies, such as the hot spots on sample "Bottle", which could lead to misclassification. 
%Moreover, FastFlow, DRAEM, and FAVAE failed to localize the anomalies in these cases, due to their modeling strategies focusing solely on normality. 
%For example, while FAVAE is capable of modeling the distribution of a single class, such as bottles, it struggles to accurately model other classes, even when trained on multiple classes. As a result, it often predicts a bottle-like shape for all classes, highlighting its limitation in handling diverse categories.

\begin{table}
\centering
\caption{In-distribution study of unsupervised anomaly detection on MVTecAD dataset. Bold font indicates the best result, and italics indicate the second-best result.}
\begin{tabular}{l|lll|lll}
\hline
\multicolumn{1}{c|}{\multirow{1}{*}{Method}} & \multicolumn{3}{c|}{AP (\%)$\uparrow$}                                                       & \multicolumn{3}{c}{AUROC (\%)$\uparrow$}                                                    \\ \cline{2-7} 
\multicolumn{1}{c|}{}                        & \multicolumn{1}{c}{Low} & \multicolumn{1}{c}{Mid} & \multicolumn{1}{c|}{High} & \multicolumn{1}{c}{Low} & \multicolumn{1}{c}{Mid} & \multicolumn{1}{c}{High} \\ \hline
FastFlow \cite{yu2021fastflow} & 0.41 & 22.5 & 5.15 & 68.4 & 67.5 & 61.2 \\
CutPaste \cite{li_cutpaste_nodate} & 0.56 & 22.5 & 6.37 & 65.5 & 66.6 & 63.6 \\
DRAEM \cite{zavrtanik_draem_2021}    & 0.23 & 30.4 & 21.8 & 11.4 & 50.4 & 62.6 \\
FAVAE  \cite{dehaene2020FAVAE}   & 0.48 & 36.4 & 11.9 & 55.4 & 80.8 & 61.2 \\
STPM \cite{wang2021student}     & 7.40 & 35.8 & 15.4 & 75.0   & 86.4 & 71.5 \\\hline 
uniAD \cite{uniAD}    & \textit{39.7} & \textit{66.1} & \textit{47.1} & \textit{99.1}   & \textbf{94.8} & \textbf{95.8} \\
InDeed & \textbf{46.7} & \textbf{74.8} & \textbf{51.8} & \textbf{99.5} & \textit{94.6} & \textit{90.0}  \\ \hline 
\end{tabular}
\label{Table:AD}
\end{table}

\begin{table}
\caption{Generalizability in terms of unsupervised anomaly detection. Symbol $\Delta$ refers to the decrease compared with the performance on the clean image.}
\label{Table: Generalization and robustness of AD}
\scalebox{0.75}{
\begin{tabular}{l|ll|llll}
\hline
% \multicolumn{1}{c|}{\multirow{2}{*}{}} & \multicolumn{2}{c|}{\textcolor{blue}{Shift from details}}    & \multicolumn{4}{c}{\textcolor{blue}{Shift from principal components}}                           \\ \cline{2-7} 
\multicolumn{1}{c|}{\multirow{2}{*}{}}                 & \multicolumn{2}{c|}{Noisy MVtecAD (Mid)} & \multicolumn{2}{c|}{Severstal}    & \multicolumn{2}{c}{MOOD} \\ \cline{2-7} 
\multicolumn{1}{c|}{}                  & AUROC $\uparrow$ ($\Delta$ $\downarrow$)          & AP $\uparrow$ ($\Delta$ $\downarrow$)            & AUROC $\uparrow$ &  \multicolumn{1}{c|}{AP $\uparrow$}    &AUROC $\uparrow$        & AP $\uparrow$         \\ \hline
DRAEM                                 & N/A                & N/A               & 40.1  & \multicolumn{1}{l|}{5.56} & N/A           & N/A        \\
FastFlow                          & N/A                & N/A               & 55.5  & \multicolumn{1}{l|}{10.7} & N/A           & N/A          \\
FAVAE                                  & N/A                & N/A               & 83.3  & \multicolumn{1}{l|}{32.1} & N/A           & N/A          \\
STPM                              & 69.4 (17.0)      & \textit{28.2} (\textit{27.4})      & 85.8  & \multicolumn{1}{l|}{37.0}   & 73.4        & 1.61       \\
uniAD                                & \textit{73.0} (\textit{21.8})      & 10.5 (55.6)     & \textbf{90.4}  & \multicolumn{1}{l|}{\textit{38.9}} & \textit{80.6}        & \textit{1.67}       \\
InDeed                                  & \textbf{90.4} (\textbf{4.22})      & \textbf{69.1} (\textbf{5.74})     & \textit{87.8}  & \multicolumn{1}{l|}{\textbf{46.2}} & \textbf{82.4}        & \textbf{44.3}       \\ \hline
\end{tabular}}
\end{table}

\textbf{OOD performance.}
We further evaluated the generalizability with three OOD datasets, \textit{i.e.}, noisy MVTecAD, Severstal, and MOOD.
Note that for the noisy MVTecAD, the performance of each method was reported in group Mid. The completed results and visualizations are presented in Section 5.2 of Supplementary Material. 
% For the complete results and visualization for comparisons, please refer to Section 5.2  in Supplementary Material.

Table~\ref{Table: Generalization and robustness of AD} provides the results. 
Overall, only STPM, uniAD, and InDeed produced reasonable results across all datasets, whereas the other three methods failed on both Noisy MVTecAD and MOOD.
For Noisy MVtecAD, InDeed achieved the best AP (69.1\%) and  AUROC (90.4\%) scores, demonstrating robustness with only a slight decrease of 5.74\% in AP and 4.22\% in AUROC, compared to the noise-free scenario.
For the Severstal and MOOD datasets, InDeed and uniAD achieved comparable AUROC, but similar to the in-distribution scenario, uniAD had evidently lower AP scores.

\subsection{Studies of OOD active generalization}
\label{Section4: Active generalization study}
\begin{figure*}
    \centering
    \includegraphics[width=0.95\linewidth]{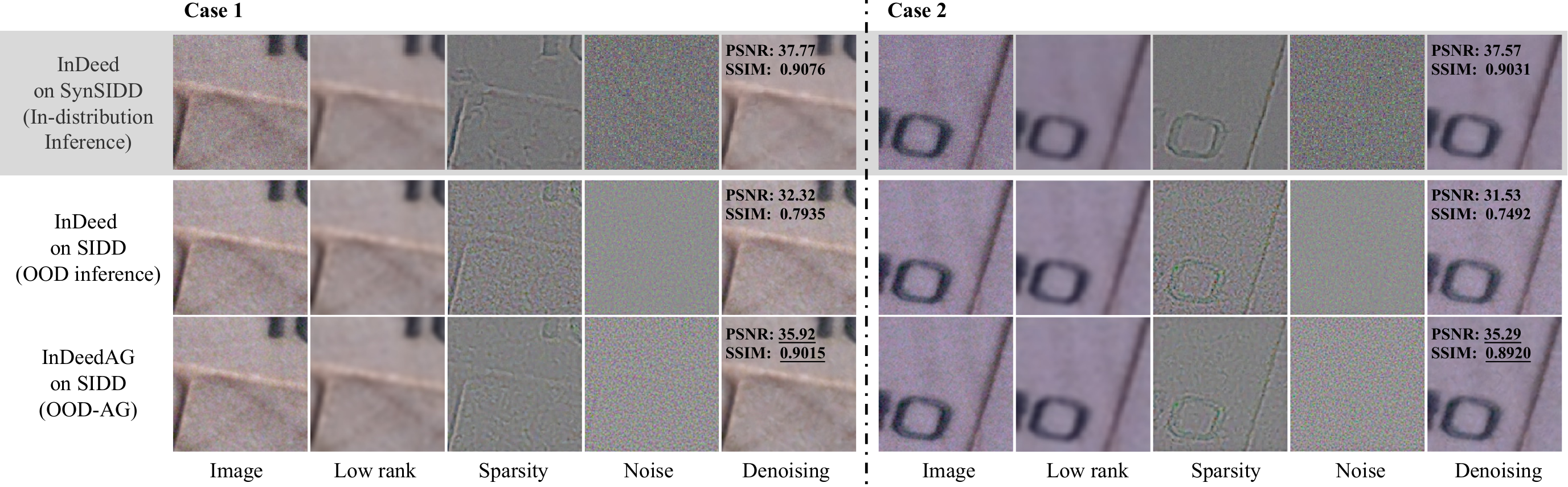}
    \caption{Visual comparisons between InDeed and InDeedAG on two cases from SIDD (cropped patches of size $100 \times 100$). Note that "InDeed on SynSIDD" serves as reference, representing predictions using synthetic in-distribution images with AWGN ($\sigma=15$). One can see that InDeedAG achieves better decomposition results than InDeed, particularly evident in noise components, leading to better PSNR and SSIM. }
    \label{Interpretable/vis_TTA_SIDD}
\end{figure*}

\begin{figure*}
     \centering
     \begin{subfigure}[b]{0.33\textwidth}
         \centering
         \includegraphics[height=5.2cm]{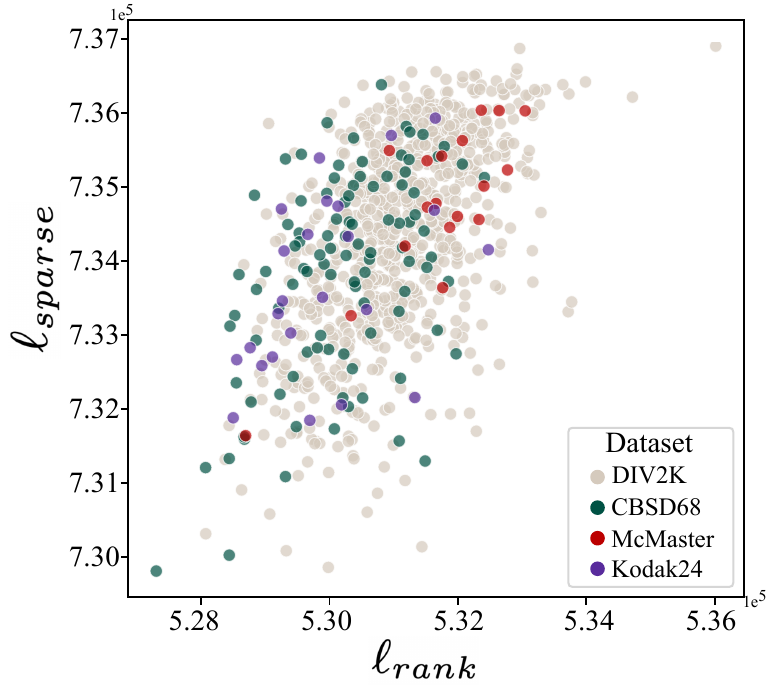}
         \captionsetup{font={footnotesize}}
         \caption{In distributions (ID)}
         \label{fig:ID data}
     \end{subfigure}
     \begin{subfigure}[b]{0.33\textwidth}
         \centering
         \includegraphics[height=5.2cm]{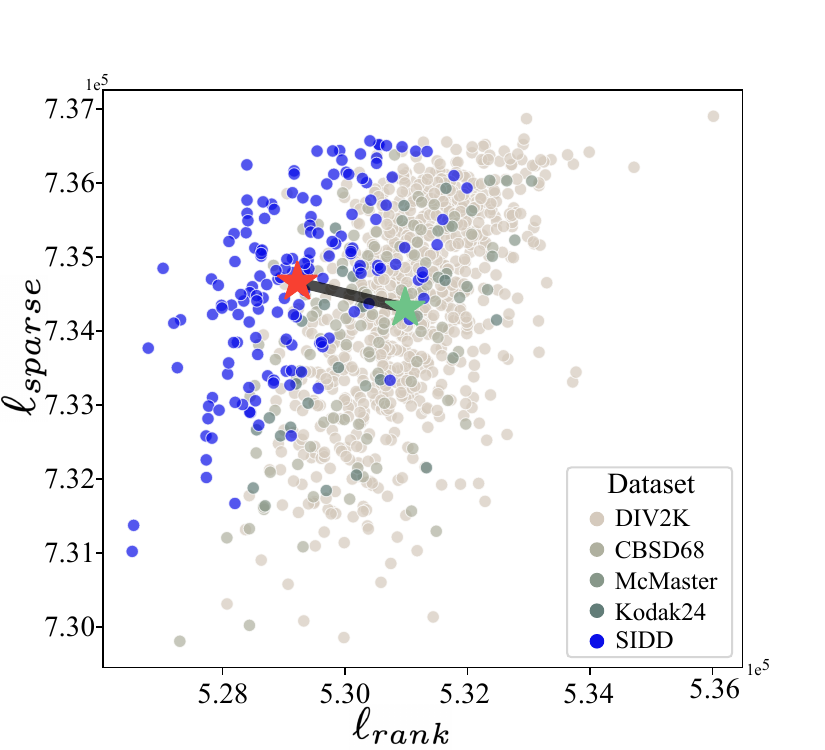}
         \captionsetup{font={footnotesize}}
         \caption{Out of distributions (OOD)}
         \label{fig:OOD data}
     \end{subfigure}
     \begin{subfigure}[b]{0.33\textwidth}
         \centering
         \includegraphics[height=5.2cm]{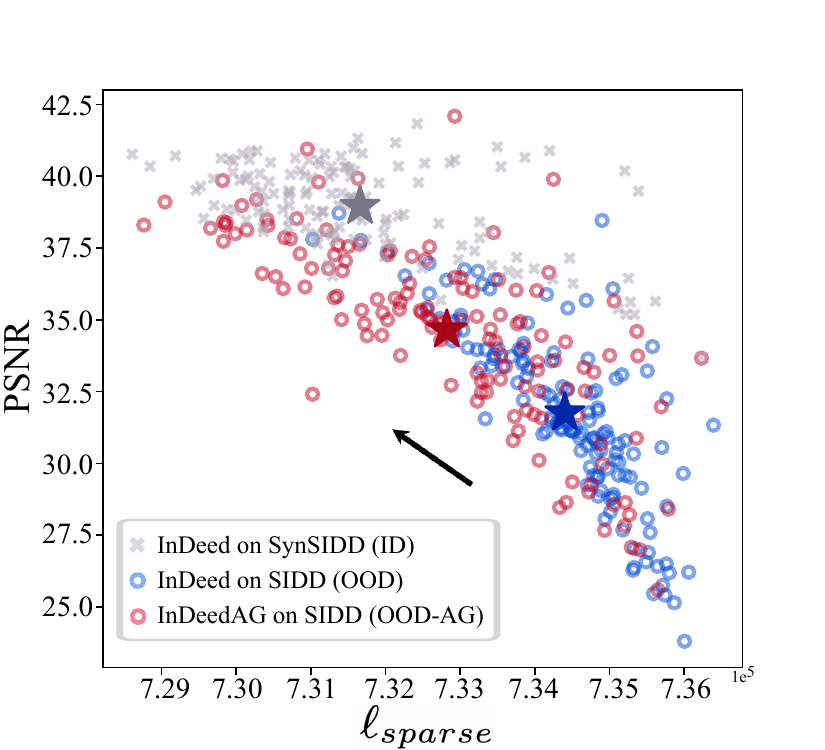}
         \captionsetup{font={footnotesize}}
         \caption{Active generalization analysis}
         \label{fig:OOD}
     \end{subfigure}
        \caption{Impacts of $\ell_{rank}$ and $\ell_{sparse}$ on identifying OOD datasets (in (a-b)) and facilitating active generalization (in (c)). (a) and (b): Joint distributions of $(\ell_{rank},\ell_{sparse})$ for in-distribution (ID) and OOD test datasets, respectively. Gray points represent training samples from DIV2K. In (b), one can see that SIDD shows OOD behavior, with its average point (red star) deviating from DIV2K's (green star). (c): Joint distributions of $(\ell_{sparse}, \mathrm{PSNR})$ for InDeed and InDeedAG. InDeed on SynSIDD represents ID inference. InDeed on SIDD shows higher $\ell_{sparse}$ and lower PSNR, whereas InDeedAG reduces loss and boosts performance. The arrow illustrates the trend from InDeed's mean point to InDeedAG's.
        % (c): Visualization of the joint distribution of  $(\ell_{rank}, \mathrm{PSNR})$ for in-distribution using InDeed on Synthesized SIDD dataset (InDeed on SynSIDD), 
        % OOD on real SIDD dataset (InDeed on SIDD),
        % and OOD on SIDD  with active generalization (InDeedAG on SIDD). 
        % In (c), InDeedAG reduces the gap between OOD and in-distribution scenarios, compared with InDeed without active.
}
        \label{Figure/Interpretable/In-distribution_OOD}
\end{figure*}

To study the effectiveness of OOD active generalization (AG), namely InDeedAG, we adaptively finetuned the pertrained InDeed model with OOD test data.
During finetuning, the learning rate $\eta$ and  batch size were set as $1 \times 10^{-6}$ and $1$, respectively.

\subsubsection{Image Denoising} 
We used the SIDD and PolyU datasets for real-world image denoising study.
Four variants of active generalization were studied:
(1) InDeedAG: solely adapt the sparsity module  using Eq.~(\ref{eq:active generalization loss}),
(2) InDeedAG$\beta$ (L): solely adapt the low-rank module using Eq.~(\ref{eq:active generalization loss:L}),
(3) InDeedAG$\beta$ (LS): adapt the two modules simultaneously using Eq.~(\ref{eq:active generalization loss:LS}),
(4) InDeedOAG: online active generalization with adaptation solely on the sparsity module for each test image.

%algorithm was applied to adapt module $f_{\theta_S}$ of InDeed, resulting in InDeedAG. We also explored two other adaptation strategies: adapting $f_{\theta_L}$ and adapting both modules, producing InDeedAG$\beta$ (L) and InDeedAG$\beta$ (LS), respectively. Moreover, we applied the online active generalization algorithm on module $f_{\theta_S}$ for each image, resulting in InDeedOAG.

\textbf{Results.} 
Table~\ref{Table: generalization for DEN} presents the results. 
InDeedAG outperformed all the OOD tested models without active generalization in Table~\ref{Table: generalization for DEN} (A).
Particularly, active generalization boosted performance of InDeed 
on SIDD by 2.61 dB in PSNR and 0.1152 in SSIM, and 
on PolyU by 1.49 dB in PSNR and 0.0311 in SSIM. 
Notably, the adaptation times (ADP) on the two datasets were only 28s and 20s, respectively, highlighting the efficiency of InDeedAG.
In addition, InDeedOAG further improved the performance of InDeedAG on SIDD with an increase of 2.19 dB in PSNR and 0.1328 in SSIM, though the improvement on PolyU was not evident. 
Again, InDeedOAG was also efficient, with 0.04-0.07 second expense per image for online active generalization. 

%It achieved the best results on both SIDD and PolyU datasets, except on the C1 and C2 sub-datasets.   The superiority of InDeedOAG came from the individual optimization for each image, differing from InDeed and InDeedAG, which were optimized on image data simultaneously. Additionally, the average adaptation times per image were just 0.07s for SIDD and 0.04s for PolyU, demonstrating the efficiency of InDeedOAG. 

Comparing the results solely in Table~\ref{Table: generalization for DEN} (B), InDeedAG, with adaptation solely on the sparsity module ($f_{\theta_S}$), performed evidently better than InDeedAG$\beta$ (L), 
demonstrating that adapting $f_{\theta_S}$ is more effective in handling noise shifts.
This is because $f_{\theta_S}$ is designed to capture sparse image details from noisy data, making it more sensitive to OOD noise. 
In contrast, the low-rank module ($f_{\theta_L}$) captures low-rank information, which nevertheless focuses on global consistency and thus is inherently robust to perturbations like noise shift.
In addition, one can see that InDeedAG$\beta$ (LS) did not show significant improvement on either dataset and even sometimes performed worse than InDeedAG$\beta$ (L). This suggests that simultaneously adapting multiple modules can pose additional challenges to the framework.

\textbf{Visualization for interpretation.} Fig.~\ref{Interpretable/vis_TTA_SIDD} shows the image decomposition results on two samples from the SIDD dataset. 
Here, \textit{OOD inference} and \textit{OOD-AG} refer to results predicted by InDeed without active generalization and InDeedAG, respectively. 
For reference, we also present the results of \textit{In-distribution inference}, in which InDeed was applied to synthetic noisy images, referred to as SynSIDD, created by adding AWGN with $\sigma=15$ to the ground truth images.
% For reference, we also presented the results of \textit{In-distribution inference},  where InDeed was utilized for synthetic noisy images, which were generated by adding AWGN ($\sigma=15$) on the ground truths from SIDD, denoted as SynSIDD.
% \textit{In-distribution inference} denotes the results produced by InDeed on synthetic noisy images, which were generated by adding AWGN ($\sigma=15$) on the ground truths from SIDD, denoted as SynSIDD. 
% \textit{Out-of-distribution inference} and \textit{OOD active generalization} refer to results predicted by InDeed and InDeedAG on SIDD, respectively. 
One can see that the predicted sparsity components ($S$) from OOD inference were more intertwined with noise, resulting in much lower values of PSNR and SSIM. 
However, after active generalization the sparsity and noise components were decomposed more separately by InDeedAG, leading to improved PSNR and SSIM on both cases.

\textbf{Interpretation in terms of loss functions.} 
We further elaborate on the effect of active generalization from the aspect of loss terms, \textit{i.e.}, $\ell_{sparse}$ and $\ell_{rank}$ in Eq.~\eqref{eq: deep loss function}. 

Firstly, to study the effect of such pair ($\ell_{sparse}$, $\ell_{rank}$) in indicating the severity of OOD, we calculated them for each image from different datasets, and visualized their distributions.
% Firstly, we calculated 
% the pair ($\ell_{sparse}$, $\ell_{rank}$) for each image from different datasets, and visualized their distributions in  Fig.~\ref{Figure/Interpretable/In-distribution_OOD} (a)-(b).
In Fig.~\ref{Figure/Interpretable/In-distribution_OOD} (a), sample points of CBSD68, McMaster, and Kodak24 with AGWN ($\sigma=15$), are distributed inside the distribution of the training dataset, \textit{i.e.,} DIV2K, confirming in-distribution (ID). 
% In contrast, in Fig.~\ref{Figure/Interpretable/In-distribution_OOD} (b) the sample points from SIDD, an OOD dataset compared with DIV2K, exhibit a distribution of lower level of $\ell_{rank}$ and larger $\ell_{sparse}$,  with its average point (red star) deviating from DIV2K’s (green star). 
In contrast, Fig.~\ref{Figure/Interpretable/In-distribution_OOD} (b) shows that sample points from SIDD, an OOD dataset compared to DIV2K, exhibit a distribution of lower level of $\ell_{rank}$ and larger $\ell_{sparse}$. One can notice the disparity from the distinct deviation of its average point (red star) from DIV2K’s (green star).
% We illustrated this di screpancy of distributions using the distance between the mean values of the two distributions, respectively marked in red (SIDD) and green (DIV2K) star symbols.
These observations are consistent with our experimental design, illustrating the shift from Gaussian noise to real-world noise,  
suggesting the pair of loss terms serve as effective metrics for assessing distribution discrepancy.
% These observations are consistent with our experimental setting, showing the loss functions pair as an appropriate indicator of distribution alignment with the training data.

Secondly, we study the effect of OOD active generalization. 
Note that the increased $\ell_{sparse}$ in OOD datasets can be a main source impeding model generalizability. 
Hence, the adaptation of $f_{\theta_S}$ could be effective. 
Fig.~\ref{Figure/Interpretable/In-distribution_OOD} (c) visualizes the distribution of the value pair ($\ell_{sparse}$, PSNR), predicted by InDeed and InDeedAG with adaptation on the sparsity module. 
Fig.~\ref{Figure/Interpretable/In-distribution_OOD} (c) also presents the distribution of InDeed on in-distribution (ID) SynSIDD as reference.
% visualizing the distribution of the value pair ($\ell_{sparse}$, PSNR), predicted by InDeed and InDeedAG on the SIDD dataset.
% Fig.~\ref{Figure/Interpretable/In-distribution_OOD} (c) 
% illustrates the effect of active generalization, by visualizing the distribution of the value pair ($\ell_{sparse}$, PSNR), predicted by InDeed and InDeedAG on the SIDD dataset. Gray points indicate results on SynSIDD.
Compared with the synthetic ID reference, InDeed on SIDD (OOD) demonstrated relatively higher $\ell_{sparse}$ and lower PSNR values. 
In contrast, after fine-tuning $f_{\theta_S}$, InDeedAG reduced $\ell_{sparse}$ significantly, resulting in higher PSNR values.

\begin{figure}
    \centering
    \includegraphics[width=1\linewidth]{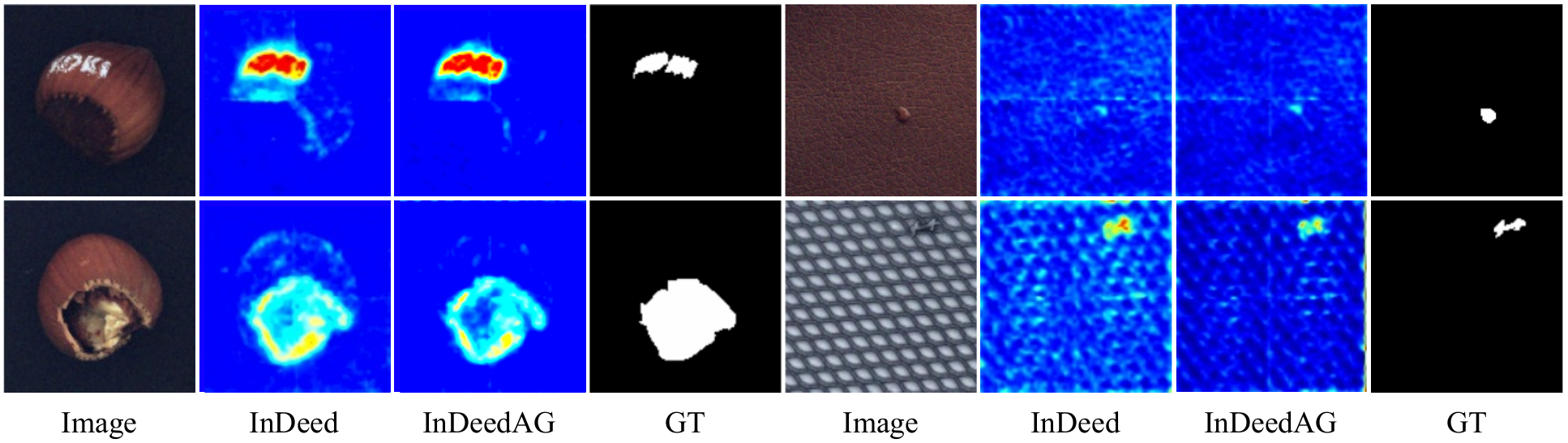}
    \caption{Visual comparisons between InDeed and InDeedAG: the latter can further removes the false positives in UAD. The color map images are the anomaly maps. GT:
ground truth segmentation map.}
    \label{Interpertable-Vis-TTA-Brain}
\end{figure}

\subsubsection{Unsupervised anomaly detection}
For this study, we adapted $f_{\theta_S}$ of InDeed to the noisy MVTecAD dataset via active generalization (InDeedAG) and online active generalization (InDeedOAG), for comparisons.
% We adopted InDeedAG, which fine-tuned the $f_{\theta_S}$ of InDeed on the noisy samples.

InDeedAG achieved AP and AUROC scores of 69.5 and 90.5, respectively, surpassing InDeed by 0.04 in AP and 0.01 in AUROC;
and InDeedOAG surpassed InDeed by 0.01 in AP and 0.02 in AUROC. 
One can read that the performance gains in terms of AP and AUROC for both of the two active generalization methods were limited. 
%, due to the complex decision boundary of $f_{\theta_S}$ for anomaly detection, which could collapse more easily.
We therefore further investigated the anomaly maps predicted by these methods. 
Fig.~\ref{Interpertable-Vis-TTA-Brain} visualizes four typical cases from InDeed and InDeedAG.
One can see that the predictions of InDeed were more crippled by noise, with blurry delineation of defects and false-positive patterns. In contrast, InDeedAG alleviated the noise effect, resulting in clearer defect boundaries and fewer false-positive pixels in the background. 
This visual investigation indicated that active generalization could further remove false positives, though the quantitative metrics may not be able to capture this performance gain.

% However, the improvement after active generalization is limited,
% due to the complex decision boundary of $f_{\theta_S}$ for anomaly detection, which more easier to collapse. 

\begin{figure*}[t]
    \centering
    \includegraphics[width=0.9\linewidth]{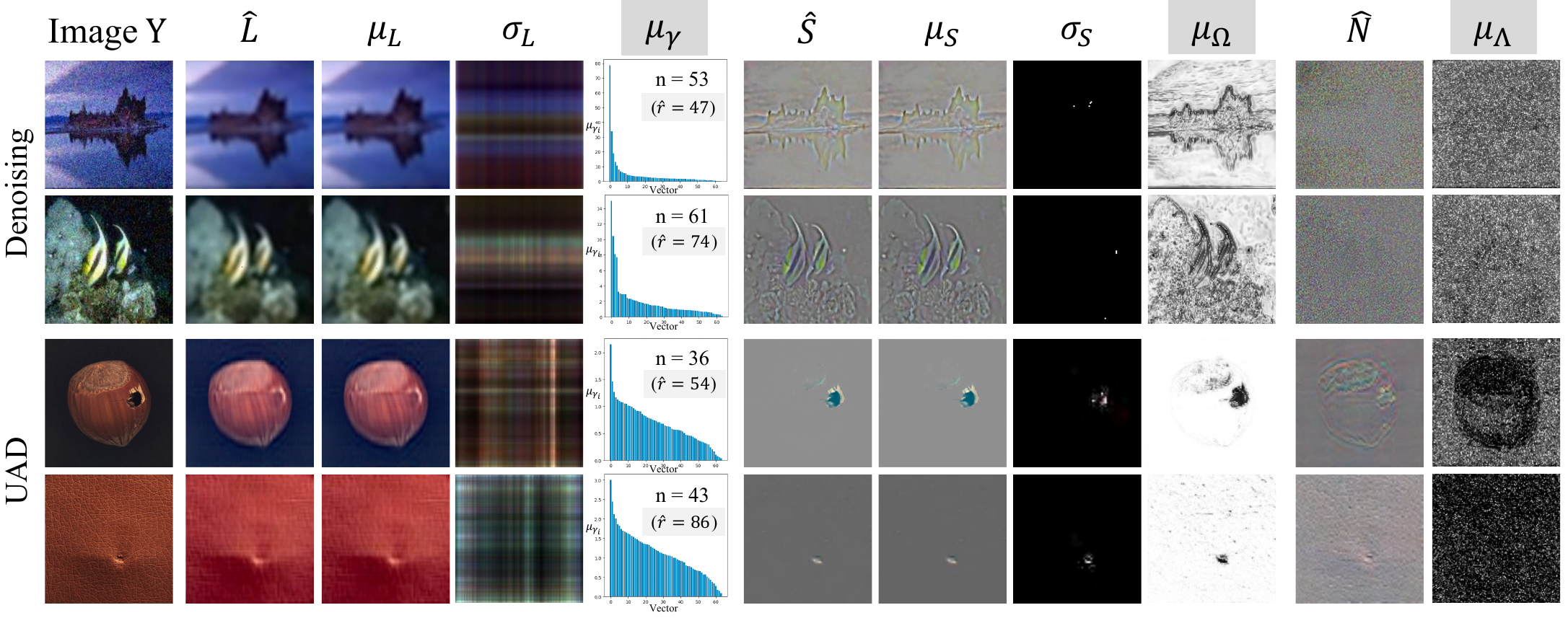}%Figure/Interpretable/Interpretable_posterior1.pdf
    \caption{Visualization of posteriors inferred by InDeed for two tasks. For middle-level variables, the sampled result ($\hat{L}/\hat{S}/\hat{N}$), mean ($\mu_{L/S}$), and standard deviation $(\sigma_{L/S}$) are shown. For leaf-level variables, expectations ($\mu_{\boldsymbol{\gamma}/\Omega/\Lambda}$) are displayed. Note that for $\sigma_{L/S}$ and $\mu_{\Omega/\Lambda}$, we display the logarithmic values, (\textit{e.g.}, $\log(\sigma_L)$) for better visual effects.
    For $\mu_{\gamma} \in \mathbb{R}^{r^0}$, we show the histogram for each image, along with the image rank $\hat{r}$ computed via SVD and the proposed index $n$ for reference.
    For $\mu_{\Omega/\Lambda}$, we calculated the norm pixel-wisely, with brighter pixels indicating a higher possibility of being zero in corresponding samples.} 
    \label{Figure/Interpretable/Interpretable_vis}
\end{figure*}

\begin{figure*}
     % \centering
     \begin{subfigure}[b]{0.2\textwidth}
         \centering
         \includegraphics[height=3cm]{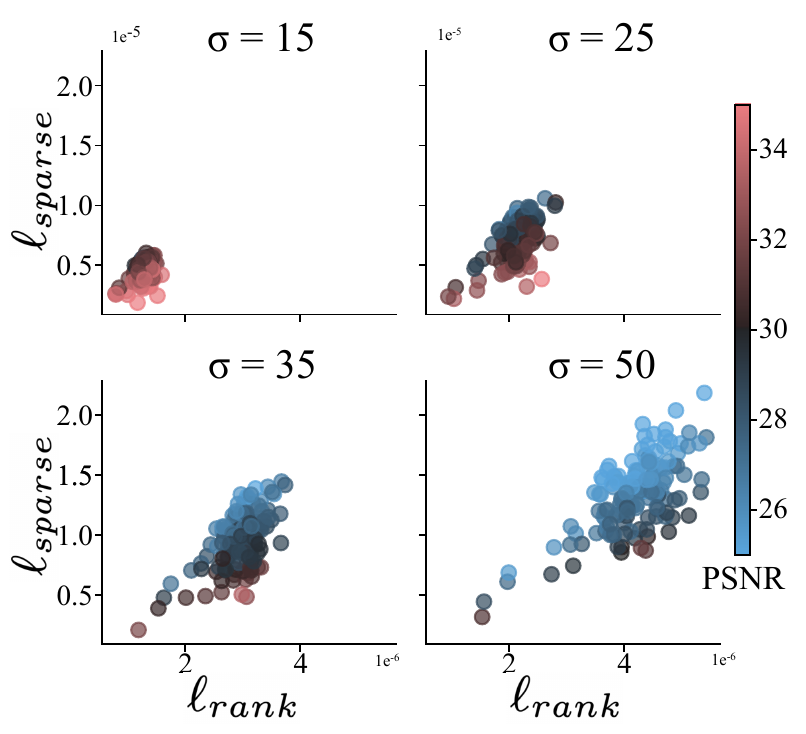}
         \caption{PSNR}
         \label{fig:PSNR vs kl_l}
     \end{subfigure}
     \begin{subfigure}[b]{0.2\textwidth}
         \centering
         \includegraphics[height=3cm]{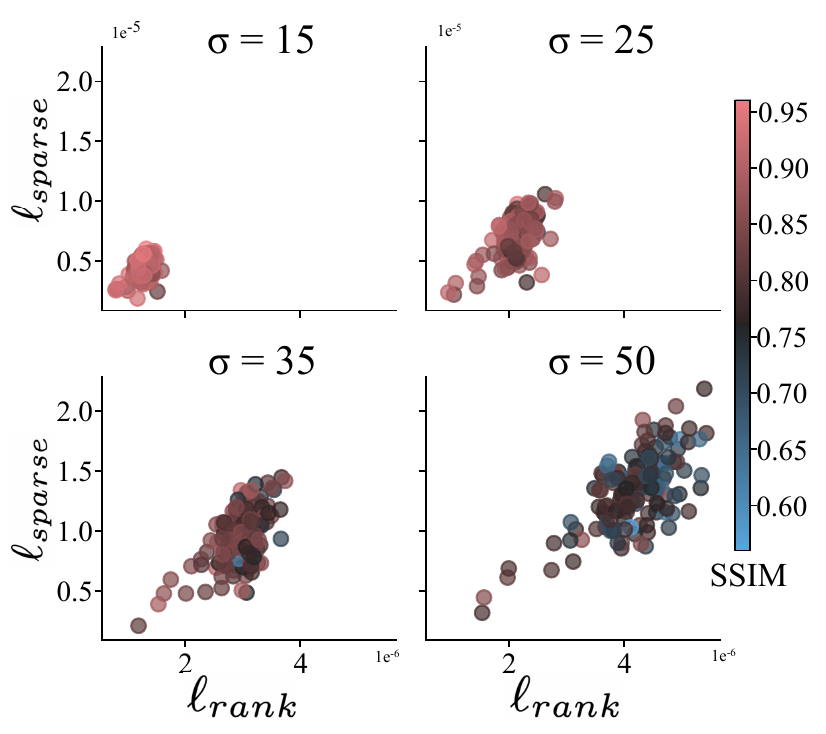}
         \caption{SSIM}
         \label{fig:SSIM vs kl_l}
     \end{subfigure}
     \begin{subfigure}[b]{0.2\textwidth}
         \centering
         \includegraphics[height=3cm]{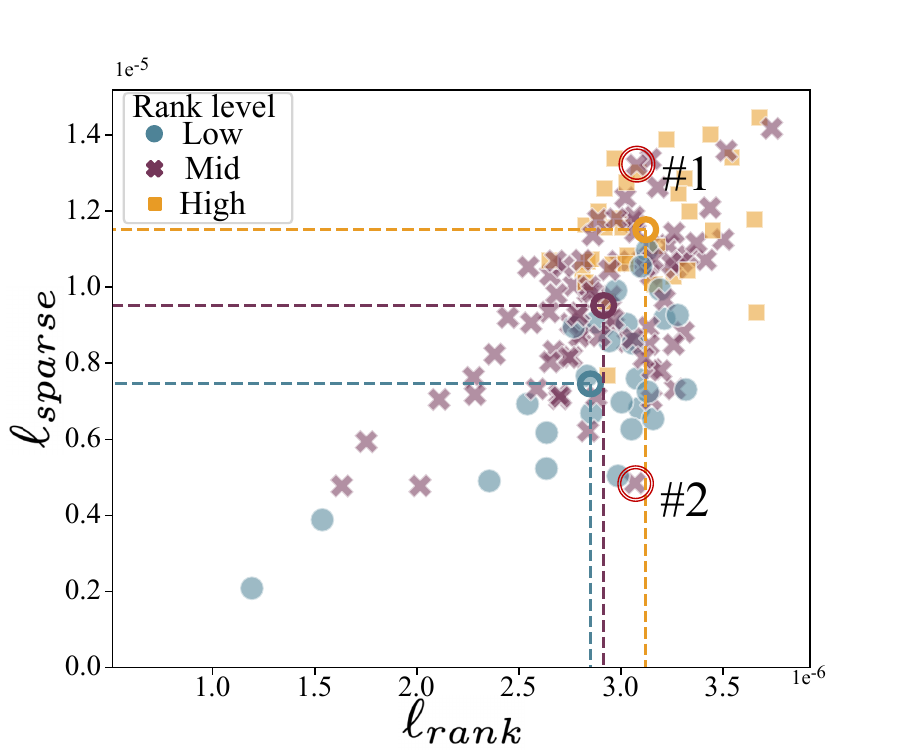}
         \caption{$(\ell_{rank}, \ell_{sparse})$}
         \label{fig:interpre_plot}
     \end{subfigure}
     \begin{subfigure}[b]{0.2\textwidth}
         \centering
         \includegraphics[height=3cm]{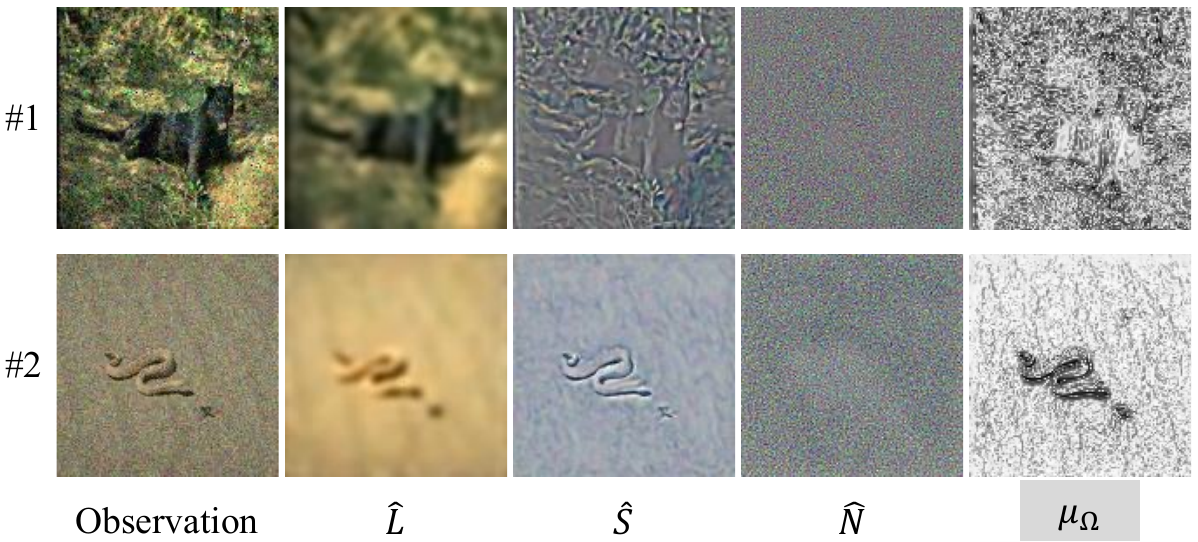}
         \caption{Visualization}
         \label{fig:interpre_vis}
     \end{subfigure}
        \caption{Relationship between $(\ell_{rank}, \ \ell_{sparse})$ and denoising performance (in (a-b)) and image details (in (c-d)). (a) and (b) show the correlation with performance, quantified by (a) PSNR and (b) SSIM, under different noise levels. Each point refers to an ID test image. (c) looks at the effect on image detail. Images are categorized by their ranks determined via SVD. One can see that the image with a higher rank tends to have a higher $\ell_{rank}$. Furthermore, despite similar $\ell_{rank}$ levels, as demonstrated in Case \#1 and \#2, $\ell_{sparse}$ can differ. (d) compares the decomposition results for Cases \#1 and \#2, illustrating that their $S$ present distinct levels in terms of the richness of details, controlled by $\mu_\Omega$ with higher values.}
        \label{Figure/Interpretable/Rank_plus_vis}
\end{figure*}

\subsection{Interpretability study}
\label{Section4:interpretability}
In this section, we investigated the interpretability of InDeed from two aspects, (1) the representation of intermediate outputs and (2) the interpretation of the low rank and sparsity terms, \textit{i.e.}, $\ell_{rank}$ and $\ell_{sparse}$ in Eq.~\eqref{eq: deep loss function}.

\subsubsection{Visualization of intermediate outputs}
To demonstrate the interpretability of each module, we visualized the outputs of InDeed, namely posteriors of variables $\mathcal{Z}= \{L, S, N, \boldsymbol{\gamma}, \Omega, \Lambda \}$, as Fig.~\ref{Figure/Interpretable/Interpretable_vis} shows. Note that for $L$,
its mean and variance can be explicitly computed from those of $A$ and $B$, \textit{i.e.,}
$\mu_{L} = \frac{\mu_A (\sigma_B^T)^2 + \sigma_A^2\mu^T_B}{\sigma^2_A I_B^T + I_A(\sigma^T_B)^2}$ and $\sigma^2_{L} = \frac{\sigma^2_A(\sigma^T_B)^2}{\sigma^2_A I_B^T + I_A(\sigma^T_B)^2}$. Moreover, for reference  we introduced an index based on $\boldsymbol{\gamma}$, defined as $n=|I|$, where $I = \{i|p(\gamma_i^{-1}>threshold)>0.95\}$.
% reflect the estimated rank for each image, 
% $I = \{i|\mu_{\gamma_i}^{-1}>0.1*\max(\mu_{\gamma}^{-1})\}$.
% or $I = \{i|p({\gamma_i}^{-1}>threshold)>0.95\}$?
% The hierarchical image decomposition modeling provides interpretations for decomposed components through variables $\mathcal{Z} = \{L, S, N, \boldsymbol{\gamma}, \Omega, \Lambda \}$.
% % enables us to decouple the image into interpretable components through variables $\mathcal{Z} = \{L, S, N, \boldsymbol{\gamma}, \Omega, \Lambda \}$, 
% The proper representation of them can lead to better model performance.
% To demonstrate this,  we visualized the inferred posteriors of these variables with samples from image denoising and unsupervised anomaly detection tasks, as shown in Fig.~\ref{Figure/Interpretable/Interpretable_vis}.

% Note that for the low-rank component $L$,
% % for better understanding, instead of the posterior of $A$ and $B$, the visualization of that of $L$ is given, whose parameters are explicitly computed via  
% its mean and variance can be explicitly computed via
% $\mu_{L} = \frac{\mu_A \sigma_B^2 + \mu_B \sigma_A^2}{\sigma^2_A + \sigma^2_B}$ and $\sigma_{L} = \frac{\sigma^2_A\sigma^2_B}{\sigma^2_A + \sigma^2_B}$. 
% One can Chat the input image $Y$ is decomposed into its low-rank $L$, sparsity $S$, and noise $N$ components. Concretely, 
For the low-rank components, we visualized a sample $\hat{L}$, its mean $\mu_{L}$, standard deviation $\sigma_L$, and the expectation of leaf-level variable $\boldsymbol{\gamma}$, namely $\mu_{\gamma}$. 
One can see that $L$ captures the structure information (dominant features) of an image, in both denoising and UAD, which is consistent with our modeling. 
Note that for UAD, $L$ approximates the low-rank representation of normal images rather than those with anomalies, which is guided by the modeling of target $U$ formulated in Eq.~\eqref{loss: AD supervision term}.
% For instance, in the hazelnut case, $\hat{L}$ attempts to represent the low-rank characteristics of a flawless hazelnut, aligning with its supervision modeling of $U$ in Eq.~\eqref{loss: AD supervision term} and benefiting from the self-supervision strategy.
Moreover, the histograms of $\mu_{\boldsymbol{\gamma}}$ shows the capability of rank adaptation for each case. 
% Indicator $n$ summarizes the behavior of $\boldsymbol{\gamma}$ and is related to the image rank $\hat{r}$ to some extent. 
% Specifically, in the image denoising task, a higher $\hat{r}$ correlates with higher $\hat{r}$ values. 
The indicator of estimated rank $n$ is close to $\hat{r}$ for image denoising in the two cases.
In the second case where $\hat{r}=74$, $n$ is close to the maximum value $r^{0}=64$, which is the upper limit of our rank modeling.
In the two cases of UAD, there is a significant gap between $n$ and $\hat{r}$.
This is desirable, because $L$ models the normal low-rank pattern, whereas $\hat{r}$ calculates the rank from the original images that contained anomalies, resulting in the inherent deviation. 
Although InDeed shows rank adaptability, accurately determining the rank of low-rank estimation remains a challenge.

For the sparsity components, we visualized a sample $\hat{S}$, mean $\mu_{S}$, standard deviation $\sigma_S$, and mean value of the sparsity-controlling variable $\Omega$, namely $\mu_{\Omega}$. 
Note that $\hat{S}$ represents task-specific sparsity information, guided by the distinct supervision losses, \textit{i.e.}, $\ell_{sup|_{\mathrm{DEN}}}$ and $\ell_{sup|_{\mathrm{UAD}}}$ in Eq.~\eqref{l_sup_DEN} and Eq.~\eqref{loss: AD supervision term}. 
For denoising, $S$ captured structural information such as boundaries, while for UAD, it extracts patterns distinct from normal regions and without high self-consistency, such as anomalies.
% S extracted
% anomalies, which had distinct patterns from the normal
% regions with high self-consistency.
% information that is sparse compared to the high self-consistency of the content, such as anomalies. 
% The results of task-specific patterns are guided by the distinct supervision losses, \textit{i.e.}, $\ell_{\mathrm{sup|_{DEN}}}$ and $\ell_{\mathrm{sup|_{UAD}}}$ in Eq.~\eqref{l_sup_DEN} and Eq.~\eqref{loss: AD supervision term}, respectively. 
% Moreover, $\mu_\Omega$ values vary among images, controlling the sparsity for each specific case. Specifically, elements with higher $\mu_\Omega$ values drive corresponding values in $S$ to zero, aligning with the modeling in Eq.~\eqref{eq: sparse modeling}. 
Moreover, the figures for $\mu_\Omega$ demonstrate high consistency with the sparsity modeling.
For example, the high values of $\mu_\Omega$ in the background of the images mean a high possibility of sparsity in these areas, which is consistent with the presentation in $\hat{S}$. 
% For example, the background areas, which have less texture, are highlighted.
% Moreover, the $S$ in denoising and UAD tasks show different sparsity patterns. 
% For denoising, $S$ captures texture information, such as boundaries, while for UAD, $S$ extracts sparse defects.

For the noise components, $N$ exhibits different representations for each task. 
For image denoising, $N$ captures the pixel-wise Gaussian noise, while for UAD,  it represents the residual texture information. 
% \textcolor{red}{When this information is incorporated into $\hat{N}$, it enhances the performance of anomaly detection with $\hat{S}$.} (wfp: I prefer deleting this explanation.)

\subsubsection{Interpretation of loss function}
We interpreted the low-rank and sparsity terms, namely, $\ell_{rank}$ and $\ell_{sparse}$ from two aspects, taking the image denoising as an example.

Firstly, we analyzed the relationship between pair $(\ell_{rank},\ell_{sparse})$ and performance. 
In the in-distribution study of image denoising in Section \ref{Section4:image denoising}, we calculated triplets $(\ell_{rank},\ell_{sparse}, \textrm{PSNR}/\textrm{SSIM})$ for all images with noise levels $\sigma \in \{15, 25, 35, 50\}$ using the trained model, and illustrated their distributions in Fig.~\ref{Figure/Interpretable/Rank_plus_vis} (a) and (b), respectively. 
We had two findings.
One is that with the increase of both $\ell_{rank}$ and $\ell_{sparse}$, the  performance of denoising decreases consistently at different noise levels. 
The other is that PSNR is more sensitive to the variation of $\ell_{sparse}$. 
The observation highlights the potential of active generalization, as Section~\ref{Section4: Active generalization study} has demonstrated.

Secondly, we further interpreted the relationship between $(\ell_{rank},\ell_{sparse})$ and image details. 
As Fig.~\ref{Figure/Interpretable/Rank_plus_vis} (c) shows, each image behaves differently in terms of the pair of loss function  $(\ell_{rank}, \ell_{sparse})$. 
As the level of image rank increases, both  $\ell_{rank}$ and $\ell_{sparse}$ rise, with  $\ell_{sparse}$ showing a more significant elevation.
This suggests that the AG algorithm be more effective in adapting $f_{\theta_S}$ via minimizing $\ell_{sparse}$.
% images with higher levels of rank tend to exhibit higher $\ell_{rank}$, which results in a relatively large KL divergence, challenging InDeed's ability to extract an optimal posterior. 
Additionally, $\ell_{sparse}$ reflects the richness of details and can vary significantly even among images with similar $\ell_{rank}$. 
For instance, samples \#1 and \#2, whose details are displayed in Fig.~\ref{Figure/Interpretable/Rank_plus_vis} (c) and (d), have similar $\ell_{rank}$ values but differ evidently in texture details. 
Specifically, sample \#1 has a more complex background, leading to a higher value of $\ell_{sparse}$,  while sample \#2 exhibits more self-consistency thus with a lower value of $\ell_{sparse}$.
\emph{Section 6 of Supplementary Material provides more discussion of interpretation. }

\section{Conclusion} 
In this work, we have proposed a novel framework for developing architecture-modularized and interpretable DNNs for image decomposition and downstream tasks. 
This framework consists of three steps: modeling image decomposition, formulating inference as two optimization problems, and designing a modularized network informed by the first two steps. 
We further investigated a generalization error bound, based on which, we proposed a test-time adaptation method for out-of-distribution (OOD) scenarios. 
Finally, we demonstrated the generalizability and interpretability with two applications, \textit{i.e.}, image denoising and unsupervised anomaly detection. 

For future work, firstly models are yet to be developed, capable of integrating different decomposition rules for various applications, such as image super-resolution and inpainting.  
Secondly, we need to develop new strategies to accommodate more scenarios, by implementing the adaptation either on low-rank module $f_{\theta_L}$ or on both modules ($f_{\theta_L}$ and $f_{\theta_S}$) effectively. 
Last but not least, it is interesting to further explore the advantages of modularization, such as module reusability across different applications for transfer learning.
%\textcolor{blue}{In future work, we aim to investigate the universality of the proposed framework for integrating diverse image-prior modeling approaches across various applications. Additionally, we will explore the versatility of the modular framework, with a focus on module adaptation, combinability, and reusability. Specifically, for module adaptation, our research will concentrate on developing optimization strategies to achieve stable adaptation in complex scenarios, including those requiring adaptation on $f_{\theta_L}$ and even simultaneous adaptations.}

\section*{Acknowledgments}
This work was funded by the National Natural Science Foundation of China (grant No. 62372115). The authors would like to thank Boming Wang, An Sui, Hangqi Zhou, and Yibo Gao for useful comments and proofreading of the manuscript.

% \section{Proof of the Second Zonklar Equation}
% Appendix two text goes here.}

% \section{References Section}
% You can use a bibliography generated by BibTeX as a .bbl file.
%  BibTeX documentation can be easily obtained at:
%  http://mirror.ctan.org/biblio/bibtex/contrib/doc/
%  The IEEEtran BibTeX style support page is:
%  http://www.michaelshell.org/tex/ieeetran/bibtex/
 
% \bibliography{IEEEabrv, references.bib}
% \bibliographystyle{IEEEtran.cls}
% \bibliography{reference}
\begin{scriptsize}
  \bibliographystyle{IEEEtran.bst}
  \bibliography{reference}
\end{scriptsize}
\vskip 0pt plus -1fil
\begin{IEEEbiography}[{\includegraphics[width=1in,height=1.25in,clip,keepaspectratio]{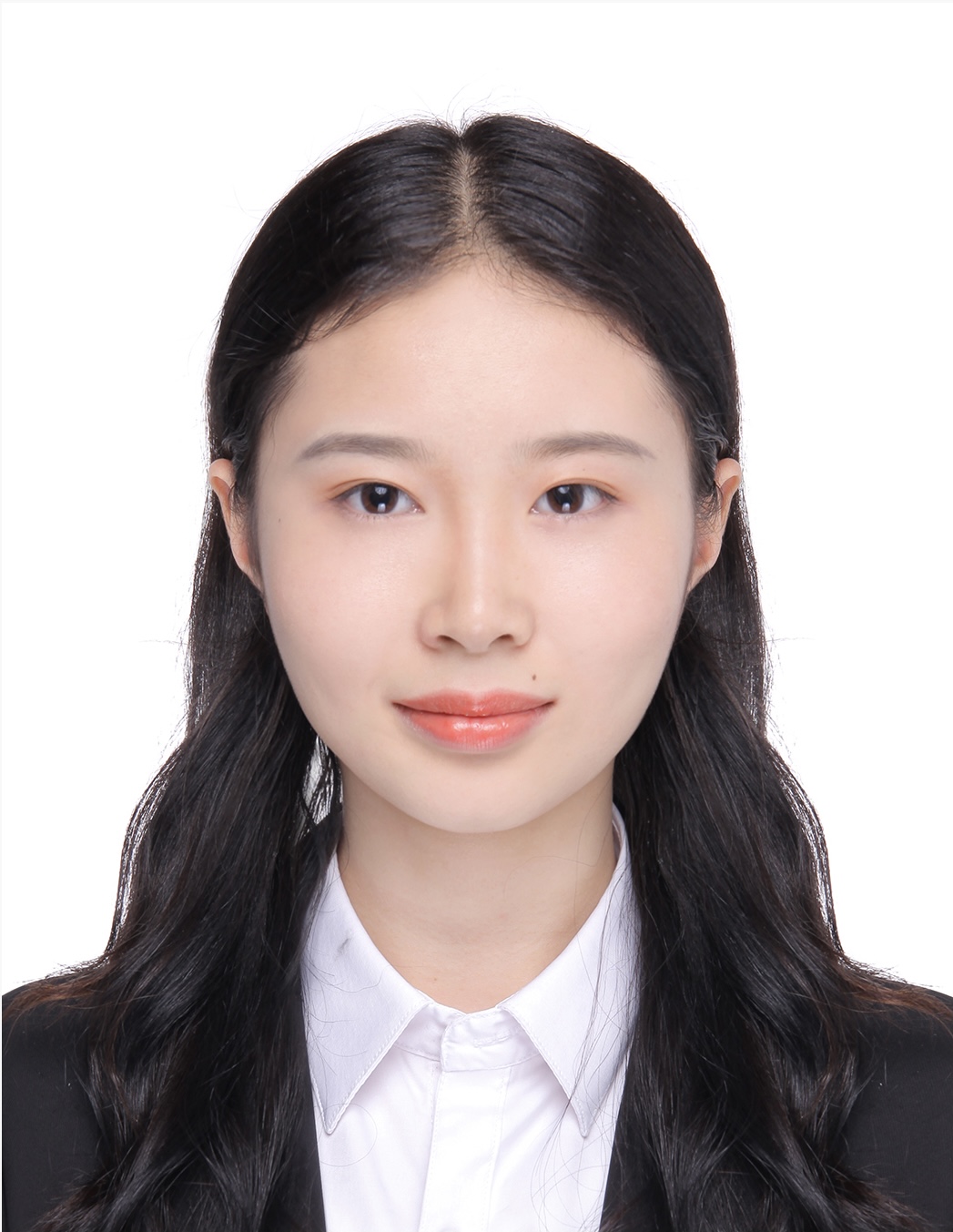}}]{Sihan Wang}
is doing a Ph.D. at the School of Data Science, Fudan University, Shanghai, China, with Prof. Xiahai Zhuang. She received her Bachelor's degree from Fudan University in 2021. Her current research interests focus on model interpretability, generalizability, and modularity for computer vision, with applications in tasks such as image restoration and segmentation.
\end{IEEEbiography}

\vspace{11pt}

\begin{IEEEbiography}[{\includegraphics[width=1in,height=1.25in,clip,keepaspectratio]{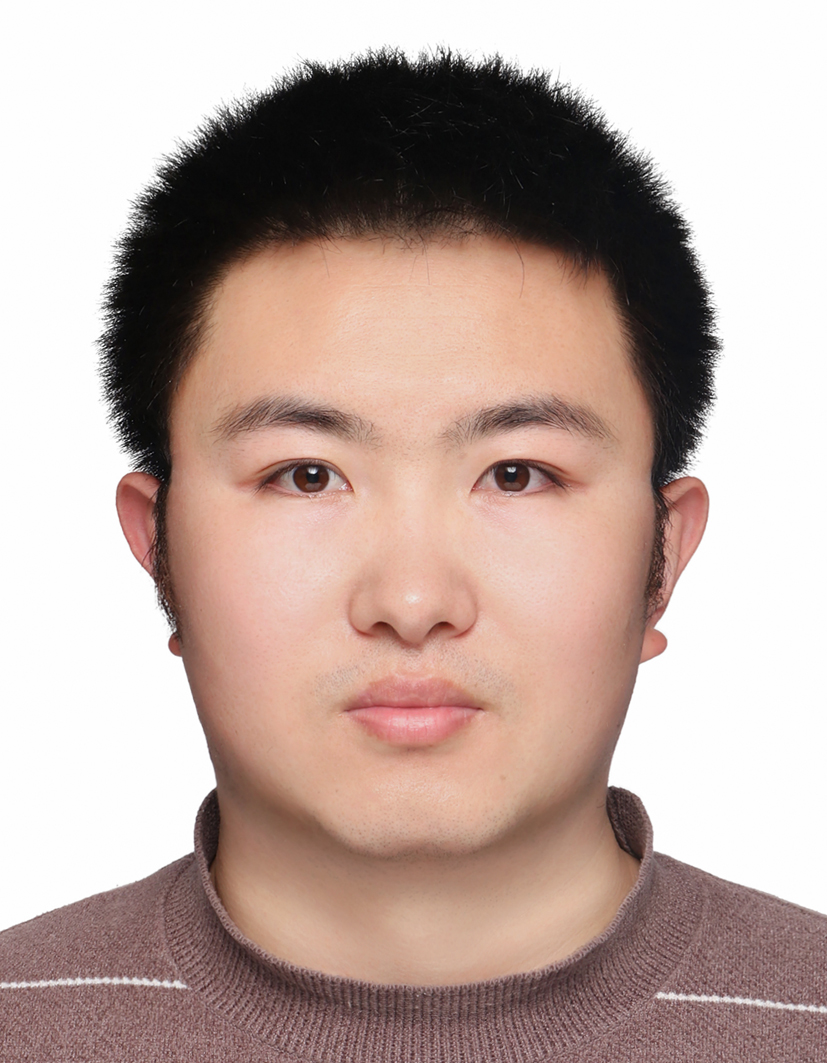}}]{Shangqi Gao} is a Research Associate at the University of Cambridge. Before that, he was a Research Assistant at the University of Oxford. He got his Ph.D. in School of Data Science, Fudan University, in 2022.
Before that he had both M.Sc and B.S. degrees in Mathematics and Statistics.
%He received the M.Sc. degree at School of Mathematics and Statistics., Wuhan University in 2018, and the B.S. degree from Faculty of Mathematics, Northwestern Polytechnical University, in 2015. 
His current research interests include computational imaging, medical image analysis and explainable AI. His work won the Elsevier-MedIA 1st Prize and Medical image Analysis MICCAI Best Paper Award 2023.
\end{IEEEbiography}

\vspace{11pt}

\begin{IEEEbiography}[{\includegraphics[width=1in,height=1.25in,clip,keepaspectratio]{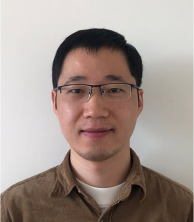}}]{Fuping Wu} is a Research Associate at the University of Oxford. He
graduated from Huazhong University of Science and Technology in 2012. He received an MSc degree from Wuhan University in 2016, and a PhD degree from Fudan University in 2021. His current research interests include medical image analysis and computer vision.
\end{IEEEbiography}

\vspace{11pt}

\begin{IEEEbiography}[{\includegraphics[width=1in,height=1.25in,clip,keepaspectratio]{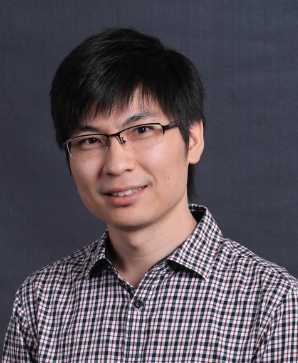}}]{Xiahai Zhuang}
is a professor at School of Data Science, Fudan University. He graduated from the Department of Computer Science, Tianjin University, received a Master's degree from Shanghai Jiao Tong University, and a Doctorate degree from University College London. His research interests include interpretable Al, medical image analysis and computer vision. His work won the Elsevier-MedIA 1st Prize and Medical image Analysis MICCAI Best Paper Award 2023.
\end{IEEEbiography}

\vspace{11pt}

\vfill

\end{document}